\documentclass[12pt]{article}

\usepackage[T1]{fontenc}
\usepackage[utf8]{inputenc}
\usepackage{authblk}
\usepackage{textcomp}
\usepackage[font=singlespacing]{caption}
\usepackage{setspace}
\usepackage{multirow}

\usepackage[english]{babel}

\usepackage{amsmath,amsthm,amssymb,amsopn,amsfonts}
\usepackage{graphics}
\usepackage{graphicx}
\usepackage{color}

\usepackage{subcaption}
\usepackage{algorithm}
\usepackage{algorithmic}

\usepackage{verbatim}
\usepackage{url}

\newcommand{\twospace}{\renewcommand{\baselinestretch}{1.3}\normalsize}

\newcommand{\R}{\mathbb{R}}

\newcommand{\frakn}{\mathfrak{n}}
\newcommand{\tr}{\textup{trace}}

\newcommand{\Dcal}{\mathcal{D}}
\newcommand{\argmax}{\arg\max}

\usepackage{mathptmx} 
\usepackage{verbatim}
\usepackage{pgfgantt}
\usepackage[normalem]{ulem}

\newtheorem{theorem}{Theorem}

\newtheorem{defn}[theorem]{Definition}

\graphicspath{{./}{./../}{./../graphs/}{./graphs/}{./sgm12figuresALL/}}
\twospace

\title{\bf Seeded Graph Matching}

\author[1]{Donniell~E.~Fishkind
\thanks{
The authors gratefully acknowledge support from the National Security Science and Engineering Faculty Fellowship (NSSEFF), the DARPA XDATA program, 
the DARPA D3M program, DARPA MAA program, and
the Acheson J. Duncan Fund for the Advancement of Research in Statistics (Awards 16-20, 16-23, 17-22, and 18-3).
Some work on this paper was undertaken at the Isaac Newton Institute for Mathematical Sciences, Cambridge, UK, during the program for Theoretical Foundations for Statistical Network Analysis,
supported by EPSRC grant no. EP/K032208/1. Part of the computation for this research project was conducted using computational resources at the Maryland Advanced Research Computing Center (MARCC).
The views and conclusions contained herein are those of the authors and should not be interpreted as necessarily representing the official policies or endorsements, either  expressed or implied, of the Air Force Research Laboratory and DARPA, or the U.S.Government.
}
}
\author[2]{Sancar~Adali}
\author[1]{Heather~G.~Patsolic}
\author[1]{Lingyao~Meng}
\author[3]{Digvijay~Singh}
\author[4]{Vince~Lyzinski}
\author[1]{Carey~E.~Priebe}

\affil[1]{Department of Applied Mathematics and Statistics, Johns
    Hopkins University}
\affil[2]{Raytheon BBN Technologies}
\affil[3]{Information Security Institute, Department of Computer Science, 
    Johns Hopkins University}
\affil[4]{Department of Mathematics and Statistics, 
    University of Massachusetts Amherst}

\begin{document}

\def\spacingset#1{\renewcommand{\baselinestretch}%
{#1}\small\normalsize} \spacingset{1}

\maketitle
\begin{abstract}
Given two graphs, the graph matching problem is to align the
two vertex sets so as to minimize the number of adjacency disagreements
between the two graphs.
The {\textit{seeded}} graph matching problem is the graph matching problem
when we are first given a partial alignment that we are tasked
with completing.
In this article, we modify the state-of-the-art approximate graph matching
algorithm ``FAQ'' of Vogelstein et al. (2015)
to make it a fast approximate seeded graph matching algorithm,
 adapt its applicability to include graphs with
differently sized vertex sets, and 
extend the algorithm so as to provide,
for each individual vertex, a nomination list of likely matches.
We demonstrate the effectiveness of our algorithm via simulation and real data experiments; indeed, knowledge of even a few seeds can be extremely effective when our seeded graph
matching algorithm is used to recover a naturally existing alignment that is only partially observed.
\end{abstract}

\noindent%

\newpage
\spacingset{1.45}

\section{Introduction}
\label{intro}


It is increasingly common in many scientific
disciplines to represent the complex interactions
amongst data points via networks, and the statistical
and scientific literature has seen a vast amount of
work on the modeling and analysis of this burgeoning
network data.  Often when working simultaneously with
multiple graphs, there is a need to register vertices
to each other across networks.  This data processing
step, which is often referred to as graph matching,
allows for the application of techniques that use the
common, shared labels across networks as known
parameters in subsequent inference.  As such, the
field of graph matching has been the subject of a
vast amount of literature in many diverse fields such
as pattern recognition \cite{lu2016fast,
riesen2016predicting, lerouge2017new}, computer
vision \cite{sang2012robust, egozi2013probabilistic,
iodice2015salient}, social network analysis
\cite{pedarsani2011privacy, yartseva2013performance},
and neuroscience \cite{chvolypr16joint,yang2009graph},
to name a few; see the following surveys for a
comprehensive overview of the many recent results,
advances, and applications of graph matching: 
``Thirty Years of Graph Matching in Pattern Recognition''
by Conte et al (2004) \cite{cofosave04},
``Graph Matching and Learning in Pattern Recognition in
the Last 10 Years'' by Foggia et al (2014) 
\cite{fopeve14},
and ``A Short Survey of Recent Advances in Graph Matching''
by Yan et al (2016) \cite{yaetal16}.

Informally, given two graphs of the same order, the graph matching problem seeks a bijection between the vertex sets of the two graphs that best preserves the adjacency structure across networks.  
More formally, suppose we are given two graphs, $G_1$ and $G_2$, with respective vertex
sets $V_1$ and $V_2$ such that $|V_1| = |V_2|$,
where $|V|$ denotes the cardinality of $V$.
For any bijective function $\phi : V_1 \rightarrow V_2$,
two vertices $u,v \in V_1$ are said to have an adjacency disagreement
under $\phi$ if $u$ and $v$ are adjacent in $G_1$ but $\phi(u)$ and
$\phi(v)$ are not adjacent in $G_2$, or vice versa.
The {\it graph matching problem} 
has as its objective the minimization of the number of adjacency disagreements induced by $\phi$
over all bijective functions $\phi : V_1 \rightarrow V_2$.
In its most general form, the graph matching problem is equivalent
to the NP-hard quadratic assignment problem \cite{sago1976}.
In fact, even the simpler problem of deciding whether there exists a graph
isomorphism, while shown to be of subexponential complexity in \cite{ba15}, 
is not known to be
solvable in polynomial time and remains intractable, in general, for large networks.
Indeed, there are no known efficient algorithms for graph matching in
general, and the consensus is that none exist.

A natural question stemming from graph matching is: What
if part of the matching is fixed? Exploration of this
{\it seeded} graph matching problem is addressed in a number of
settings.
The authors of \cite{zaslavskiy2009global} and \cite{type_constraints}
incorporate constraints that enforce correspondences to be only
between vertices of the same ``type,''
and more recent advances include:
use in semisupervised learning, spectral clustering, and seed-selection processes
(see \cite{ham2005semisupervised, lyzinski2015spectral, lica15} respectively),
exploration of other methods for using seeds
in graph matching
(see \cite{hurugu13, kahagr15}),
along with a few surveys covering seeded graph matching
(see, for example, \cite{fopeve14, yaetal16}).

In the notation introduced above, the seeded graph matching problem can be formulated as follows.  
Suppose that we are also given subsets $W_1 \subset V_1$, $W_2 \subset V_2$
such that $|W_1| = |W_2|$,
and a fixed bijection $\psi: W_1 \rightarrow W_2$.
The {\it seeded graph matching problem} is defined to be the problem of
minimizing the number of adjacency disagreements induced by $\phi$
over all bijections $\phi : V_1 \rightarrow V_2$
that are {\it extensions} of $\psi$ --- that is, $\phi$ must agree with $\psi$ on $W_1$ (i.e., for all $u \in W_1$, $\phi(u)=\psi(u)$).
The elements of $W_1$ and $W_2$ are called {\it seeds} and
$\psi$ is called a {\it seeding}.

In many applications when considering two networks, there is a natural
correspondence between the two vertex sets; for
example, when looking at an email network and a social network with the
same participants, we might say that a vertex in the email network corresponds
to a vertex in the social network if the two represent the same
participant.
In this case, a goal 
of graph matching would be to recover this
natural correspondence if it is unknown.
We can therefore consider the seeded graph matching problem as an attempt to
recover an underlying vertex correspondence when only a few of these
correspondences (seeds with the seeding function) are known a priori.
Knowledge of these seeds can yield a dramatic improvement in
approximating the natural correspondence,
as will be illustrated later in this article.

The purpose of this article is to modify the
Fast Approximate Quadratic Assignment Problem (\texttt{FAQ})
graph matching algorithm of 
Vogelstein et al (2015) 
\cite{vo_etal15}, so as to
\begin{enumerate}
\item Employ the use of seeds --- we call this
the Seeded Graph Matching Algorithm ``\texttt{SGM}'', and
\item Adapt the use of the \texttt{SGM} algorithm to
match graphs which have differently sized vertex sets,
and
\item Extend the \texttt{SGM} algorithm to provide each vertex with a probability distribution over potential matches.
\end{enumerate}
The modification of \texttt{FAQ} to use seeds (``1'' above)
was first introduced in the technical report which
was an early version of this
article. 
As such,  this article has already been a stepping stone from which numerous
articles have already been written, all citing this article,
including \cite{lyadvopapr14, lyfipr14, 
lyzinski2015spectral, 
lysutaatpr14perfectase, 
filypachpr15, 
chvolypr16joint, 
lyzinski2016information,
lyfifivoprsa15, 
lylefipr16vn3}.

The motivation for specifically extending the \texttt{FAQ} algorithm
comes from the fact that the algorithm is both fast and accurate,
and also has
strong theoretical justification.
Indeed, in 
Vogelstein et al. (2015) 
\cite{vo_etal15} it is shown that
the \texttt{FAQ} algorithm
is more accurate than the other graph matching algorithms with
which it was compared on $94\%$ of the QAPLIB \cite{burkard1997qaplib} benchmark problems considered.
Also, \texttt{FAQ} was faster than the popular
PATH algorithm introduced in \cite{zaslavskiy2009path, liqixu12}.
Furthermore, from a theoretical standpoint,
it is shown in \cite{lyfifivoprsa15} that, under mild conditions, the
\texttt{FAQ} algorithm asymptotically almost surely provides the optimal graph
matching
solution when solved exactly, in contrast to many of the existing
alternative problem formulations/solutions.

As mentioned in ``2'' above, another purpose of this manuscript is to
include an extension of graph matching to settings where the vertex sets
differ in size; here the goal is to recover a correspondence between the
vertices of the smaller graph and a subset of the vertices of the larger graph.
Another purpose of this manuscript (``3'' above) is to provide,
for each vertex, a probability distribution over potential matches.  
This allows (among other things) for
ranking of possible matches,
rather than just providing a single proposed match.
In particular,
by providing each vertex with a ranked list of candidate
matching vertices, 
a practitioner is given a principled means with which to search for the true match for a given vertex, namely, to search the rank list in decreasing likelihood of matching.

In particular, our \texttt{SGM} algorithm would be appropriate to use
when two graphs are isomorphic and we seek the isomorphism, and also
when two graphs are not isomorphic, and we seek the bijection between
the their respective vertex sets that is ``closest'' to an isomorphism.
Furthermore, by Contribution ``2'' listed above, SGM would be
appropriate to  use when the graphs have differently sized vertex sets,
say $G_1$ has fewer vertices than $G_2$, and we seek the one-to-one
injection from the vertex set of $G_1$ into the vertex set of $G_2$
which provides the ``closest'' isomorphic-to-$G_1$ induced subgraph of $G_2$.
All of these tasks could be used to find a natural, underlying
correspondence between the vertex sets.  Also, by Contribution ``3''
listed above, SGM would be appropriate to use when we don't just want a
single bijection between the vertex sets, but seek backup, alternative
possibilities for natural correspondence to each vertex.

The structure of this paper is as follows:
In Section \ref{sec:notation} we provide the mathematical framework
and notation.
In Section \ref{mfo}, we describe the graph matching algorithm
\texttt{FAQ} of Vogelstein et al (2015) 
\cite{vo_etal15} and adapt it into our \texttt{SGM}
algorithm for seeded graph matching.
We then present
an extension of the \texttt{SGM} algorithm for use on graphs having differently
sized vertex sets in Section \ref{sec:SGMdiff} and,
in Section \ref{sec:softsgm}, we present a version of the \texttt{SGM}
algorithm which outputs, for each pair of nodes in $V_1 \times V_2$,
a confidence of being a match across the two graphs.
Following, in Section \ref{demos} we demonstrate the effectiveness of
the \texttt{SGM} algorithm via three real data experiments, and in
Section 4 we compare \texttt{SGM} to a seeded version of the PATH
algorithm for graph matching.  We conclude in Section \ref{disc} with a
discussion of implications and future work.


\subsection{Notation and Mathematical Framework}
\label{sec:notation}

For simplicity, all graphs in this manuscript are simple;
that is, the edges are undirected, and
there are no loops or multiple edges.
(However, of note to the reader, all of our results and algorithms
can easily be extended to the matching of directed, multi-edged,
and/or loopy graphs.)
Let $n$ and $m$ be positive integers and $\frakn=n+m$.
For notational simplicity,
let $G_1$ and $G_2$ be graphs with vertex sets
$V_1, V_2 = \{1,2,\ldots, \frakn\}$.
We will take seed sets $W_1, W_2 = \{1,\ldots,m\}$
with seeding $\psi: W_1\rightarrow W_2$ as the identity function.
(When $m=0$ we have the (unseeded) graph matching problem.)
Let $A,B \in \R^{\frakn\times \frakn}$ be the adjacency matrices
for $G_1$ and $G_2$, respectively; this means that
for all $i,j \in \{1,2,\ldots,\frakn\}$
it holds that $a_{ij}=1$ or $0$ according as vertices $i,j \in V_1$ are
adjacent in $G_1$ or not, and  $b_{ij}=1$ or $0$
according as $i,j \in V_2$ are adjacent in $G_2$ or not.
It will be useful to let $A$ and $B$ be partitioned as
\[  A =\left [
\begin{array}{cc} A_{11} & A_{12} \\ A_{21} & A_{22} \end{array} \right ]
\ \ \ \ \ \ \ \ \ B =\left [
\begin{array}{cc} B_{11} & B_{12} \\ B_{21} & B_{22} \end{array} \right ]
\]
where $A_{11},B_{11}\in \R^{m \times m}$,
$A_{12},B_{12}\in \R^{m \times n}$, $A_{21},B_{21}\in \R^{n \times m}$, and
$A_{22},B_{22}\in \R^{n \times n}$.
The seeded graph matching problem can be expressed as
\[
    \min\limits_{P\in \Pi_n}
    \|A-(I_{m}\oplus P)B(I_{m}\oplus P)^T\|_F^2,
\]
where $\Pi_n$ denotes the set of $n \times n$ permutation matrices,
$I_{m}$ is the $m$-by-$m$
identity matrix,
$\oplus$ is the direct sum of matrices,
and $\| \cdot \|_F$ is the Frobenius norm on matrices.
For a given permutation matrix $P$, the corresponding bijection
$\phi_{P}: \{1,2,\ldots,\frakn\} \rightarrow \{1,2,\ldots,\frakn\}$
is defined as,
for all $i,j \in \{1,2,\ldots,\frakn\}$, $\phi_{P} (i)=j$
precisely when $[I_{m} \oplus P]_{ij}=1$.

Often, the aim of graph matching is to uncover a natural underlying correspondence
function $\Psi : V_1 \to V_2$ between entities represented by $V_1$ and $V_2$.
In order to model a setting where vertices in $G_1$ naturally correspond to
vertices in $G_2$ via $\Psi$, in our simulations of Sections 
\ref{ssec:time},
\ref{sec:sgmex}, \ref{sec:SGMdiff}, 
and \ref{sec:softsgm}, we take
$G_1$ and $G_2$ to be realizations
from a $\rho$-correlated Stochastic Block Model, (or ``SBM'').
The SBM model is introduced in \cite{Holland1983},
and the $\rho$-SBM described below is used in \cite{lyzinski2015spectral}.

\begin{defn}
    \label{def:rhosbm}
    Suppose that we are given: number of vertices $\frakn$,
    number of \textup{blocks} $k\in\mathbb{Z}>0$,
    vertex set $V = \{1,2,\ldots, \frakn\}$,
    \textup{edge probability matrix} $\Lambda \in [0,1]^{k\times k}$,
    \textup{block membership function} $b : V \to \{1,\ldots,k\},$ and
    correlation $\rho \in [0, 1]$.
    Random graphs $G_1$ and $G_2$, each having vertex set $V$
    with respective adjacency matrices $A$ and $B$, have
    a $\rho$\textup{-correlated Stochastic Block Model} distribution
    with parameters $k, b,$ and $\Lambda$,
    denoted $(G_1,G_2)\sim \rho\text{-SBM}(k,b,\Lambda)$, if
    \begin{enumerate}
        \item Each of $G_1$ and $G_2$ is marginally distributed as a
            Stochastic Block Model with parameters $k,b,\Lambda$,
            denoted $G_1,G_2\sim\text{SBM}(k,b,\Lambda)$.
                That is, for all
                    $i<j \in \{1,\ldots,\frakn\}$,
                    $A_{ij} \stackrel{ind.}{\sim}
                    \text{Bernoulli}(\Lambda_{b(i),b(j)})$ and
                    $B_{ij} \stackrel{ind.}{\sim}
                    \text{Bernoulli}(\Lambda_{b(i),b(j)})$;
        \item For all $i<j$ and $k<\ell$, the random variables
            $A_{ij}$ and $B_{k\ell}$ have Pearson correlation
            coefficient $\rho$ if $i=k$ and $j=\ell$, and are otherwise
            collectively independent.
    \end{enumerate}
\end{defn}
\noindent The last sentence in Definition \ref{def:rhosbm}
imbues $(G_1,G_2)$ with a natural vertex correspondence
between graphs, the identity map.
In this setting, $\rho=0$ implies that edge presence is independent
across the two random graphs and
$\rho=1$ implies that the two graphs are almost surely isomorphic via
identity.
Note that the $\rho$-correlated Erd\H{o}s R\`enyi Model
``$\rho$-ER($p$)'' of \cite{lyfipr14}
is a special case of the $\rho$-SBM($k,b,\Lambda$)
in which the number of blocks is effectively $k=1$,
and the more general $\rho$-correlated Bernoulli random
graph model ``$\rho$-Bernoulli($\Lambda$)'' of \cite{lyfifivoprsa15} is a special case of the
$\rho$-SBM($k,b,\Lambda$) in which the number of blocks
is $k=\frakn$.

In addition to providing a rich model for exploring the effectiveness of our \texttt{SGM} algorithm in recovering an unknown underlying correspondence, 
sampling $(G_1,G_2)$ from the $\rho\text{-SBM}(k,b,\Lambda)$ model can be easily achieved.
First, sample $G_1\sim\text{SBM}(k,b,\Lambda)$;
then, conditioned on $G_1$, sample
$G_2$ as follows: letting $A$ and $B$ denote the adjacency
matrices for $G_1$ and $G_2$, respectively, independently
sample, for $i<j$, the $ij$th element of $B$ via
$B_{ij} \sim$Bernoulli$((1-\rho)\Lambda_{b(i),b(j)} + \rho A_{ij})$.

\section{Seeded-FAQ for approximate seeded graph matching}
\label{mfo}

Since the seeded graph matching problem is intractable,
we seek an approximate solution that can be efficiently computed.
To this end,
in Section \ref{relaxation}
we express the seeded graph matching
problem as an optimization problem with integrality constraints, and then
relax the integrality constraints by replacing them with nonnegativity
constraints. In Section \ref{mfaq} we modify the \texttt{FAQ} algorithm of
Vogelstein et al (2015) 
\cite{vo_etal15} into an algorithm called
\texttt{SGM}
that approximately solves the seeded graph matching problem;
it first approximately solves the relaxed problem and then projects the
solution to restore integrality.
Following, in Section \ref{sec:SGMdiff} we generalize \texttt{SGM}
to include the matching of graphs on differently sized vertex sets, and
in Section \ref{sec:softsgm}
we extend \texttt{SGM} to create for each vertex a probability distribution over
likely matches.

\subsection{The relaxation}
\label{relaxation}

Suppose $G_1,G_2$ are graphs with respective adjacency matrices
$A, B \in \R^{\frakn\times\frakn}$, as described in
Section~\ref{sec:notation}.
As mentioned in Section \ref{sec:notation},
the seeded graph matching problem can be formulated as
\begin{equation}
    \label{eqn:defnobj}
    \min\limits_{P\in \Pi_n}
    \|A-(I_{m}\oplus P)B(I_{m}\oplus P)^T\|_F^2.
\end{equation}
Expanding, we have
\begin{equation}
    \|A-(I_{m}\oplus P)B(I_{m}\oplus P)^T\|_F^2=
\|A\|_F^2 + \|B\|_F^2
- 2 \cdot \tr A^T(I_{m}\oplus P)B(I_{m}\oplus P^T),
\end{equation}
from which we see that this optimization problem is equivalent to
\footnote{
    Note that, although $A$ and $B$ are
symmetric matrices, we nonetheless keep transposes in place wherever
they are present to enable further generalization;
our analysis will not change if we instead were
in a broader setting where $A$ and $B$ are
generic (nonsymmetric, nonhollow, and/or nonintegral) matrices in
$\R^{\frakn\times\frakn}$.
}
\begin{equation}
    \label{eqn:sgmobj}
    \max_{P\in\Pi_{n} }
    \tr \left( A^T(I_{m}\oplus P)B(I_{m}\oplus P^T)\right).
\end{equation}

To approximately solve the seeded graph matching problem it will be
useful to first relax the feasible region from $\Pi_n$,
the set of permutation matrices,
to the set of $n\times n$ doubly stochastic matrices, $\Dcal_n$;
by the Birkhoff-Von Neumann Theorem, $\Dcal_n$ is the convex hull of $\Pi_n$.
Recall that a doubly stochastic matrix is a non-negative matrix such
that all row sums and column sums equal $1$.
Thus, the relaxed seeded graph matching problem becomes
\begin{equation}
    \label{eqn:sgmrelaxobj}
    \max_{P\in\Dcal_{n} }
    \tr \left(A^T(I_{m}\oplus P)B(I_{m}\oplus P^T)\right).
\end{equation}
Indeed, this is a relaxation of seeded graph matching in the sense that
if we were to add integrality constraints --- that $P$ is
integer-valued --- then we would precisely return to the constraint
that $P$ is a permutation matrix, hence we would have
returned to the seeded graph matching problem.
The relaxed problem formulated in Equation (\ref{eqn:sgmrelaxobj}) 
is a quadratic program with an indefinite Hessian,
and thus cannot be efficiently solved exactly. In the next session we will obtain an
approximate solution using Frank-Wolfe methodology.

\subsection{From \texttt{FAQ} to \texttt{SGM}} \label{mfaq}

The \texttt{SGM} algorithm is a modification of the state-of-the-art
graph matching algorithm, \texttt{FAQ}, of 
Vogelstein et al (2015) 
\cite{vo_etal15} to allow for the use of seeds.    
\texttt{SGM} first approximately solves the relaxed seeded graph matching
problem --- maximize $\tr \left(A^T(I_{m}\oplus P)B(I_{m}\oplus P^T)\right)$
subject to $P$ being a doubly stochastic matrix --- by using the
Frank-Wolfe Method \cite{frwo56}, an iterative procedure that
involves successively solving linearizations of the quadratic objective function.  
These linearizations, it turns out,
can be here cast as linear assignment problems
that can be efficiently solved with the Hungarian Algorithm of \cite{ku55}.    
At the conclusion of Frank-Wolfe, the doubly stochastic solution obtained is projected back to
the set of permutation matrices; note that this projection step can again be cast as a linear assignment problem solvable via the Hungarian Algorithm.



We first briefly review the Frank-Wolfe Method before proceeding
to apply it. The general kind of
 optimization problem for which the Frank-Wolfe Method is used is
  \begin{eqnarray} \label{i} \textup{(FWP)  \ \ \ maximize}
\ \ \  f(x) \ \textup{   such that   } \
x \in S,
\end{eqnarray}
where $S$ is a polyhedral set (i.e., is described by linear
constraints) in a Euclidean space,
and the function $f:S \rightarrow \R$ is continuously differentiable.
A starting point $x^{(0)} \in S$ is chosen in some fashion,
perhaps arbitrarily. For $i=1,2,3,\ldots$, the following is done.
The function $\tilde{f}^{(i-1)}:S \rightarrow \R$ is defined to be the
first order (i.e., linear) approximation to $f$ at $x^{(i-1)}$ --- that is,
$\tilde{f}^{(i-1)}(x):= f(x^{(i-1)})+\nabla f(x^{(i-1)})^T(x-x^{(i-1)})$;
then solve the linear program: maximize $\tilde{f}^{(i-1)}(x)$ such that $x \in S$.  
This can be done efficiently since $\tilde{f}^{(i-1)}(x)$ is a linear
objective function with linear constraints, and note that, by ignoring additive constants, the objective function of this subproblem can be
abbreviated as: maximize $\nabla f(x^{(i-1)})^Tx$
such that $x \in S$.
Given a solution
$\tilde{x}^{(i-1)} \in S$ to this linear approximation, the point $x^{(i)} \in S$ is defined
as the solution to: maximize $f(x)$ such that $x$ is on the line segment
from $x^{(i-1)}$ to $\tilde{x}^{(i-1)}$ in $S$. 
This is a just a one dimensional
optimization problem; in the case where $f$ is quadratic this can
be exactly solved analytically. 
Go to the next $i$,
and terminate this iterative procedure
when the sequence of iterates $x^{(0)}$, $x^{(1)}$, $x^{(2)}$, \ldots stops
changing beyond a predefined threshold or develops a gradient close enough to zero. 

We now describe how \texttt{SGM} employs the Frank-Wolfe Method to
solve the relaxed  seeded graph matching problem.
The objective function to be maximized in Eq. (\ref{eqn:sgmobj}) here is
\begin{eqnarray*}  f(P)  & =  &   \tr \left (
\left [  \begin{array}{cc}  A^T_{11} & A^T_{21} \\ A^T_{12} & A^T_{22}  \end{array} \right ]
\left [  \begin{array}{cc}  I_{m} & 0_{m \times n}
\\ 0_{n \times m} & P  \end{array} \right ]
\left [  \begin{array}{cc}  B_{11} & B_{12} \\ B_{21} & B_{22}  \end{array} \right ]
\left [  \begin{array}{cc}  I_{m} & 0_{m \times n}
\\ 0_{n \times m} & P^T  \end{array} \right ]
\right ) \\
& = & \tr \left (
\left [  \begin{array}{cc}  A^T_{11} & A^T_{21} \\ A^T_{12} & A^T_{22}  \end{array} \right ]
\left [  \begin{array}{cc}  B_{11} & B_{12}P^T \\ PB_{21} & PB_{22}P^T  \end{array} \right ]
\right )\\
& = & \tr (A_{11}^TB_{11})+ \tr( A_{21}^TPB_{21})+\tr(A_{12}^TB_{12}P^T)
+ \tr(A_{22}^TPB_{22}P^T) \\
& = &  \tr(A_{11}^TB_{11})+ \tr(P^T A_{21}B_{21}^T)+\tr(P^TA_{12}^TB_{12})
+ \tr(A_{22}^TPB_{22}P^T)
\end{eqnarray*}
which has gradient
\begin{eqnarray}
    \label{eqn:gradf}
\nabla f(P):=A_{21}B_{21}^T+A_{12}^TB_{12}+A_{22}PB_{22}^T+A_{22}^TPB_{22} .
\end{eqnarray}

We start the Frank-Wolfe Algorithm at the barycenter doubly stochastic matrix
$\tilde{P}=\left(1/n\right)\vec{1}_n \vec{1}_n^T$, unless otherwise
specified, where $\vec{1}_n$ denotes the $n$-vector of all $1$'s.
In the next paragraph we describe a single step in the
Frank-Wolfe algorithm. Such steps are repeated iteratively until
the iterates empirically converge or a certain pre-selected,
fixed bound (we use 20 iterations in our examples) 
on the number of iterations is reached. 

Given any particular doubly stochastic matrix
$\tilde{P} \in \R^{n \times n}$ at which the Frank-Wolfe algorithm
currently resides,
the Frank-Wolfe-step linearization involves maximizing
$\tr(Q^T \nabla f(\tilde{P}))$ over all of the doubly
stochastic matrices $Q \in \Dcal_{n}$.
This is precisely the linear assignment problem
(since it is not hard to show that
the optimal doubly stochastic $Q$ can in fact be selected to be
a permutation matrix)
and so the Hungarian Algorithm
will in fact find the optimal $Q$, call it
$\tilde{Q}$, in $O(n^3)$ time, as shown in \cite{budema2012}.

The next task in the Frank-Wolfe algorithm step
will be maximizing the objective function over the line
segment from $\tilde{P}$ to $\tilde{Q}$;  i.e., maximizing $g(\alpha):=f(\alpha \tilde{P}
+(1-\alpha ) \tilde{Q})$ over $\alpha \in [0,1]$. Denote
$c:=\tr( A^T_{22} \tilde{P} B_{22} \tilde{P}^T)$ and
$d:=\tr (A^T_{22} \tilde{P} B_{22} \tilde{Q}^T +
    A^T_{22} \tilde{Q} B_{22} \tilde{P}^T)$ and
$e:=\tr (A^T_{22} \tilde{Q} B_{22} \tilde{Q}^T)$ and \\
$u:=\tr ( \tilde{P}^TA_{21}B_{21}^T   + \tilde{P}^TA_{12}^TB_{12} )$ and
$v:=\tr ( \tilde{Q}^TA_{21}B_{21}^T   + \tilde{Q}^TA_{12}^TB_{12} )$. Then
(ignoring the additive constant $\tr (A_{11}^TB_{11})$ without loss of
generality, since it will not affect the maximization)
we have $g(\alpha)=c \alpha^2+d \alpha (1-\alpha)
+e(1-\alpha)^2+u \alpha + v(1-\alpha)$  which simplifies to
$g(\alpha)=(c-d+e)\alpha^2+(d-2e+u-v)\alpha + (e+v)$. Setting the
derivative of $g$ to zero yields potential critical point
$\tilde{\alpha}:=\frac{-(d-2e+u-v)}{2(c-d+e)}$ (if indeed
$0 \leq \tilde{\alpha}\leq 1$); thus the next Frank-Wolfe algorithm
iterate will either be $\tilde{P}$ (in which case the algorithm would halt)
or $\tilde{Q}$ or $\tilde{\alpha}\tilde{P}+(1-\tilde{\alpha})\tilde{Q}$, and
the objective functions can be compared to decide which of these three matrices
will be the $\tilde{P}$ of the next Frank-Wolfe iterate.

At the termination of the Frank-Wolfe Algorithm, we have an
approximate solution, say $\tilde{P} \in \Dcal_n$,
to the problem maximize $\tr (A^T(I_{m}\oplus P)B(I_{m}\oplus P^T))$
subject to $P\in \Dcal_n$.

We then find the permutation matrix $\tilde{Q}$ which solves the
optimization problem min $\| Q - \tilde{P} \|_F^2$ subject to $Q \in \Pi_n$, 
and finally $\phi_{\tilde{Q}}$ is our approximate
solution to the seeded graph matching problem.
This minimization of $\|Q-\tilde{P}\|_F^2$ is equivalent
to maximizing $\tr (Q^T \tilde{P})$ subject to $Q \in \Pi_n$,
where the latter is,
again, a formulation of the linear assignment problem
solvable
with 
the Hungarian Algorithm \cite{budema2012}.

See Algorithm \ref{alg:sgm} for pseudocode of \texttt{SGM}.
\begin{algorithm}[t!]
\begin{algorithmic}
\STATE \textbf{Input}:
Graphs $G_1, G_2$, each with vertex set $\{1,\ldots,m+n\}$,
with respective adjacency matrices $A,B$, \\
\indent \indent assuming vertices $\{1,\ldots,m\}$ are seeds.\\
\textbf{Initialize}: Select $P^{(0)} \in \Dcal_{n}$, which is
$P^{(0)}:=\frac{1}{n} \vec{1}_n\vec{1}_n^T$ unless otherwise specified\\

\FOR{$i=1,2,3,\ldots$, $\langle$while stopping criteria not met$\rangle$}
\STATE \textbf{Step 1}:
Compute $\nabla f(P^{(i-1)}) = A_{21}B_{21}^T + A_{12}^T B_{12} +
A_{22}P^{(i-1)} B_{22}^T + A_{22}^T P^{(i-1)} B_{22}$;

\STATE \textbf{Step 2}:
Compute $Q^{(i-1)} \in \arg\max\tr (Q^T \nabla
f(P^{(i-1)}))$ over $Q \in\Dcal_{n}$ via the Hungarian Algorithm;

\STATE \textbf{Step 3}:
Compute step size
$\alpha^{(i-1)} \in\argmax f(\alpha P^{(i-1)} + (1-\alpha)Q^{(i-1)})$
over $\alpha \in[0,1]$;

\STATE \textbf{Step 4}:
Set next iterate
$P^{(i)} := \alpha^{(i-1)} P^{(i-1)} + (1-\alpha^{(i-1)}) Q^{(i-1)}$;
\ENDFOR

\STATE \textbf{Step 5}:
Compute $\tilde{Q} \in \arg\max \tr(Q^T P^{(i)})$ over $Q \in \Pi_n$ via the
Hungarian Algorithm;

\STATE \textbf{return} $\phi_{\tilde{Q}}$.
\end{algorithmic}
\caption{\texttt{SGM}}
\label{alg:sgm}
\end{algorithm}

\subsection{Time and space complexity of \texttt{SGM}}
\label{ssec:time}

Regarding the time complexity of the SGM algorithm, the primary computational
expenses are the Hungarian algorithm, which runs in $O(n^3)$ time
\cite{budema2012}, where $n$ is the number of non-seed vertices in the graphs,
and matrix multiplication, which naively runs in $O(n^3)$ time (assuming that
the number of seeds $m$ is $O(n)$), although there are faster implementations
of matrix multiplication \cite{coppersmith1987matrix}. Thus, given a
pre-selected maximum number of Frank-Wolfe iterates, SGM approximately solves
the graph matching problem in $O(n^3)$ time.

\begin{figure}
  \centering
  \includegraphics[width = 0.5\textwidth]{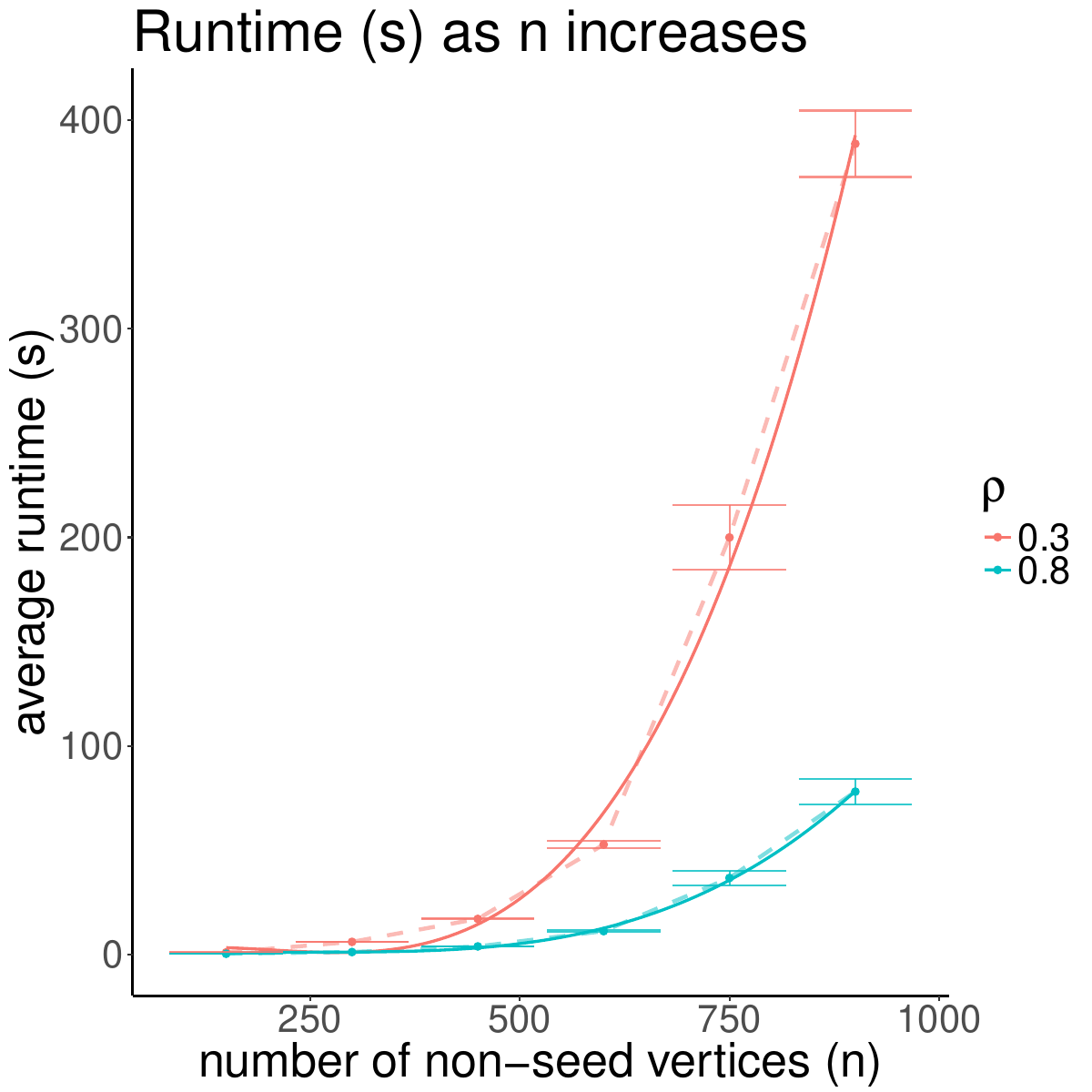}
  \caption{The average runtime (sec) plotted against the number of
    non-seed vertices $n$}
  \label{fig:time}
\end{figure}

Unless the number of seeds is excessive, they have little practical impact on
\texttt{SGM} runtime, since the input to the Hungarian algorithm is an $n
\times n$ matrix---regardless of the number of seeds---and, regarding matrix
multiplication, matrices $A_{21}B_{21}^T$ and $A_{12}^TB_{12}$ can be computed
just once at the beginning of the algorithm, and they are $n \times n$
matrices---regardless of the number of seeds---so the number of seeds has no
subsequent impact on the runtime.  Even the one computation of $A_{21}B_{21}^T$
and $A_{12}^TB_{12}$ requires relatively little time if there aren't an
excessive number of seeds. The runtime of \texttt{SGM} is thus basically a
function of $n$ and the correlation of the graphs $G_1$ and $G_2$. For each
number of non-seeds $n= 150,300,\dots, 900$, and each correlation value
$\rho=0.3,0.8$, we ran the \texttt{SGM} algorithm on 25 random instantiations
of the model used in Section \ref{sec:sgmex}, with $9$ seeds; three seeds
randomly selected from each of the three blocks. The average runtimes are
illustrated in Figure \ref{fig:time}; the runtimes indeed appear to be approximately cubic in $n$.
(Our computations were performed on Maryland Advanced Research Computing Center's system (MARCC) using standard compute nodes
with Intel Haswell dual socket 12-core processors @ 2.5GHz.)   

The space demands of \texttt{SGM} are modest. Besides the obviously
needed storage of $A$ and $B$ which, in general, requires $\Theta (n^2)$
space, the Frank-Wolfe steps do not require any history beyond the
current doubly stochastic matrix $\tilde{P}$ and its associated gradient
(which is computed by matrix multiplication). The Hungarian Algorithm
also requires just $O(n^2)$ space, as does the subsequent one
dimensional optimization (since there are three critical points to be
tested, with a fixed number of matrix multiplications). Thus, the space
requirements of SGM are just $O(n^2)$, in total. (Note that even if the
graphs are sparse, the gradients used in the Frank-Wolfe steps will not
be sparse, in general, so that sparse implementations of the Hungarian
Algorithm will not be~helpful.)

\subsection{Effectiveness of the \texttt{SGM} algorithm}
\label{sec:sgmex}

Our main inference task is to recover an underlying, existing
correspondence between the vertices of the two graphs ---utilizing
seeded graph matching, and we will soon focus on using the
\texttt{SGM} algorithm for this purpose.

However, our first simulation experiment is instead narrowly focussed 
on the optimization task, which is to check if
the \texttt{SGM} algorithm achieves the global optimum of the seeded graph
matching problem that it approximates/tries to solve. As mentioned, there is no efficient algorithm
for solving seeded graph matching, and \texttt{SGM} is an efficient algorithm,
thus we have no hope of \texttt{SGM} consistently achieving the global optimum;
indeed, in general, the most we can hope to obtain from \texttt{SGM} is
a good approximate solution. 
However, we found that  
\texttt{SGM} does an excellent job of finding the global optimum for instances where 
we can compute the global optimum (with great effort!).  
Indeed, for small graphs we can compute the
global optimum for seeded graph matching by expressing the problem as a linear integer
programming problem, and then finding the exact solution using the
discrete optimization solver {\it GUROBI  Optimizer} \cite{gurobi}. 
We independently realized $50$ pairs of
graphs $G_1,G_2$ from a $.9$-SBM($3,b,\Lambda$) distribution with
$\frakn = 30$ vertices, block assignment function $b$ such that each
of the $k=3$ blocks has $10$ vertices, $m=2$ seeds discrete-uniformly randomly
selected from the $30$ vertices, and
\begin{equation}
    \label{eqn:lambda}
    \Lambda = \begin{bmatrix}
        0.7 & 0.3 & 0.4 \\
        0.3 & 0.7 & 0.3 \\
        0.4 & 0.3 & 0.7
    \end{bmatrix}.
\end{equation}
Running the \texttt{SGM} algorithm ---as well as exactly solving
seeded graph matching via GUROBI--- on these
$50$ pairs of graphs that we instantiated, we found that all $50$ were solved
to global optimality by  \texttt{SGM}. By contrast, when we repeated all
of the above except without any seeds---in which case \texttt{SGM} is
just \texttt{FAQ}---we
found that only $23$ of the $50$ instantiations had \texttt{FAQ} achieving the global optimum.

Next, we return to the main inferential task, the estimation of an
underlying correspondence.
To demonstrate the effectiveness of the \texttt{SGM} algorithm,
we independently realized $150$ pairs of graphs $G_1,G_2$
for each value of
$m = 0,1,\ldots, 20$ and each value of $\rho = 0,0.1,0.2,\ldots,1$,
on $\frakn=n+m=300$ vertices from a
$\rho$-SBM($k,b,\Lambda$) distribution,
where $k=3$, $b$ is such that $\frakn/3 = 100$ vertices belonged to each
of the three blocks, and $\Lambda$ is as defined in Equation (\ref{eqn:lambda}).
The specific $m$ seeds used in each simulation were uniformly
randomly selected from among the $\frakn$ vertices.

For each pair of graphs, $G_1,G_2$, we ran the \texttt{SGM} algorithm, and
measured performance of the \texttt{SGM} algorithm
in terms of the {\it match ratio},
\begin{equation}
    \label{eqn:deltam}
\delta:=\frac{|\{ v \in V_1 \backslash W_1: \phi(v)=\Psi(v)\}|}{n},
\end{equation}
that is, the fraction of unseeded vertices of $G_1$ that are correctly
matched. For any particular values of $m$ and $\rho$,
let $\overline{\delta}$ denote the average match ratio taken over the 150 
realizations.

\begin{figure}[ht]
\centering
\includegraphics[scale=0.5]{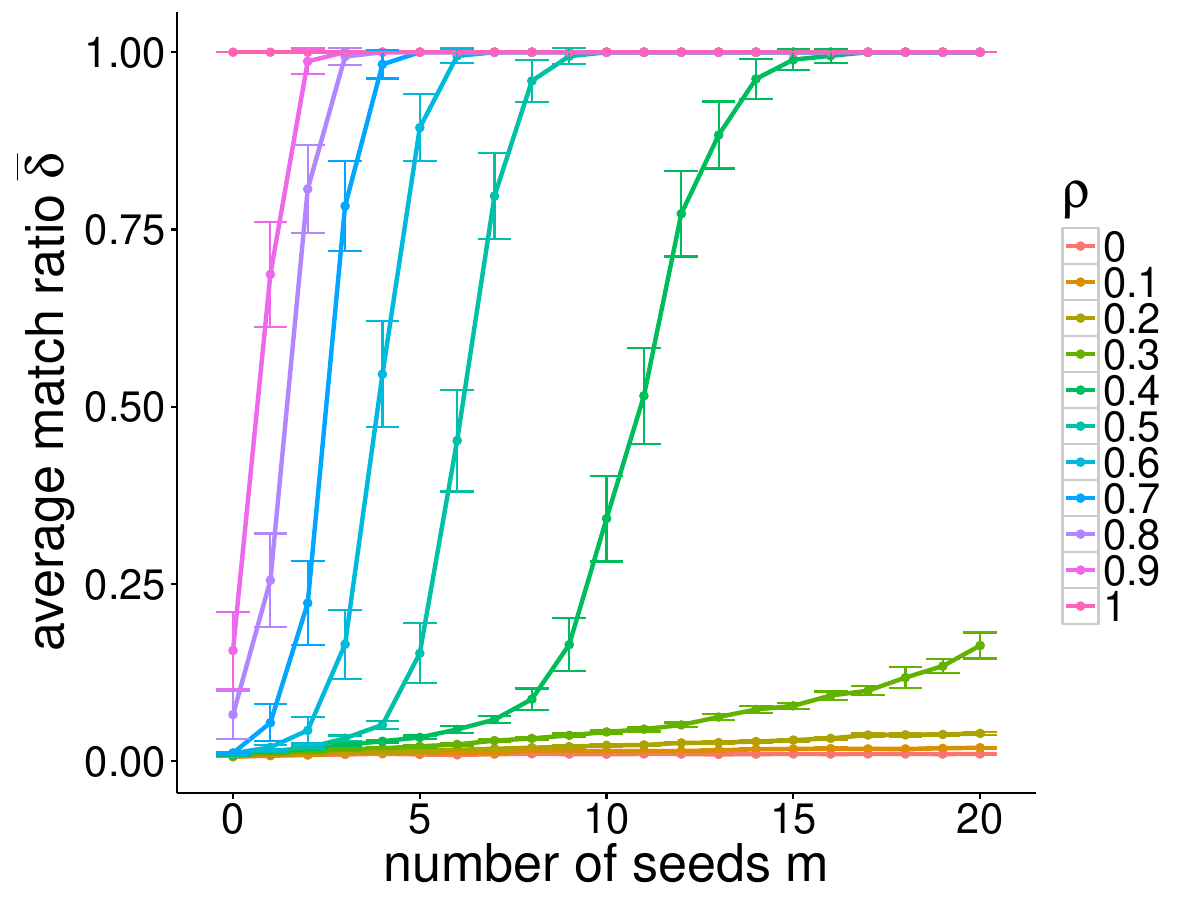}
\vspace*{-0.15in}
\caption{
    Average match ratio $\overline{\delta}\pm2$s.e.\@ as a function of the number
    of seeds $m$, for different correlation values $\rho$, in the
    $\rho$-SBM simulations of Section \ref{sec:sgmex} on $\frakn = 300$ vertices.
\label{fig:mrhovary}}
\end{figure}

For each $\rho = 0,0.1,0.2,\ldots,1.0$,
we plot in Figure \ref{fig:mrhovary} the average match ratio,
$\overline{\delta}$, along with a confidence interval,
as a function of
the number of seeds, $m=0,1,\ldots,20$.
(All confidence intervals used in this manuscript are of
the form: mean $\pm$ twice the standard error, unless otherwise specified.)
The expected number of vertices for which a discrete-uniformly randomly chosen bijection
$V_1 \rightarrow V_2$ agrees with $\Psi$ is $1$;
thus, for a chance bijection, the mean of $\overline{\delta}$ would be
$1/n = 1/(\frakn-m)$.
As expected, $\overline{\delta}$ increases as $m$ increases and also
as $\rho$ increases.
Perfect performance when $\rho=1$
indicates that the \texttt{SGM} algorithm finds the isomorphism between
graphs when it exists, while $\overline{\delta}\approx 1/n =$chance when
$\rho=0$, since the natural alignment is then meaningless.

\subsection{Graph matching when the vertex sets are different sizes}
\label{sec:SGMdiff}

Until this section, the vertex sets $V_1=\{1,2,\ldots,\mathfrak{n}\}$ and
$V_2=\{ 1,2,\ldots,\mathfrak{n}\}$ have been assumed to be
of the same cardinality as, indeed, there is assumed
to be a natural correspondence between them in the form of a
bijection $\Psi :V_1 \rightarrow V_2$. For this section, suppose that
$V_1=\{1,2,\ldots,\mathfrak{n}_1\}$ and
$V_2=\{ 1,2,\ldots,\mathfrak{n}_2\}$ are such that $\mathfrak{n}_1<\mathfrak{n}_2$,
and the underlying correspondence $\Psi :V_1 \rightarrow V_2$ is merely
injective (one-to-one), rather than bijective. We will use the expression
``core vertices'' for $V_1$ and its image
through $\Psi$; the other $\mathfrak{n}_2-\mathfrak{n}_1$
vertices in $V_2$ will be called ``extraneous vertices."
As before, the seeds are taken as $W_1,W_2 = \{1,2,\ldots,m\}$ with the
identity seeding function, for some $m < \frakn_1$

The most straightforward way to treat this seeded graph matching setting
is to pad $G_1$ in some fashion with additional vertices to bring the number of
vertices associated with $G_1$ up~to~$\mathfrak{n}_2$, and then
apply \texttt{SGM} as before.
The vertices in $V_2$ matched to the padding are then identified as the extraneous vertices, and
the remainder of the matching would approximate the natural correspondence $\Psi$ between $G_1$ and the matching subgraph of $G_2$.
It would further seem that an innocuous choice of padding is to take the
adjacency matrix $A$ for $G_1$ and append zeros to make the new
adjacency matrix $A \oplus 0_{(\mathfrak{n}_2-\mathfrak{n}_1)\times
(\mathfrak{n}_2-\mathfrak{n}_1)}$;
this consists of adding $\mathfrak{n}_2-\mathfrak{n}_1$ isolated vertices to $G_1$.

Unfortunately, this choice is not innocuous. The effect of this padding scheme is to match $G_1$ to the best fitting \emph{subgraph} of $G_2$, as opposed to the desired best fitting \emph{induced} subgraph.
Indeed, these isolated vertices of the padding
will have an affinity to be matched to low-density subgraphs in $G_2$,
even if these vertices
in $G_2$ are core vertices.
In this case, the correct correspondences
for these $G_2$ vertices will not be correctly recovered by the matching.
Indeed, in this manner
the isolated vertices of the padding carry a lot of false signal---and are not merely
the absence of signal that is desired to promote the (default) matching of the padded vertices
in $G_1$ to the extraneous vertices of $G_2$.

Going back to the previous situation where $G_1$ and $G_2$ each have $\mathfrak{n}$ vertices
and respective adjacency matrices $A,B$, define $\widetilde{A}:=2A- \vec{1}_{\mathfrak{n}}
\vec{1}_{\mathfrak{n}}^T$ and $\widetilde{B}:=2B- \vec{1}_{\mathfrak{n}}
\vec{1}_{\mathfrak{n}}^T$; these are effectively adjacency matrices, except that
$1$ or $-1$ indicate adjacency or non-adjacency, respectively,
instead of the usual $1$~or~$0$.
It is clear that substituting $\widetilde{A}$ and $\widetilde{B}$ in
place of $A$ and $B$, respectively,
in the seeded graph matching formulation of Equation (\ref{eqn:defnobj}) yields an
equivalent optimization problem.
It is then immediate that 
substituting $\widetilde{A}$ and
$\widetilde{B}$ in place of $A$ and $B$, respectively, into the trace formulation of Equation (\ref{eqn:sgmobj}) yields an
equivalent optimization problem as well.

However, in our present setting where $G_1$ and $G_2$ respectively have $\mathfrak{n}_1$ and
$\mathfrak{n}_2$ vertices and $\mathfrak{n}_1<\mathfrak{n}_2$, let us consider the
effect of substituting $\widetilde{A}\oplus 0_{(\mathfrak{n}_2-\mathfrak{n}_1)\times
(\mathfrak{n}_2-\mathfrak{n}_1)}$ and $\widetilde{B}$ in place of $A$ and $B$,
respectively, in the seeded graph matching formulation in Equation
(\ref{eqn:sgmobj}).
Indeed, for any particular permutation matrix $I_m \oplus P$ in that expression, consider
the associated matching.
The padded vertices in $G_1$ and the vertices which they
are matched to in $G_2$ collectively add $0$ to the objective function, and the objective
function is just $2$ times the difference between the number of edge agreements
and the number of edge disagreements across the matching between the original
vertices of $G_1$ and the vertices they are matched to in $G_2$.
In summary, substituting $\widetilde{A}\oplus 0_{(\mathfrak{n}_2-\mathfrak{n}_1)\times
(\mathfrak{n}_2-\mathfrak{n}_1)}$ and $\widetilde{B}$ in place of $A$ and $B$,
respectively, in the seeded graph matching formulation in Equation
(\ref{eqn:sgmobj}) is
equivalent to minimizing the number of edge disagreements under bijection
$\phi$ from the vertices of $G_1$ to the vertices of an \emph{induced subgraph} of
$G_2$ with $\mathfrak{n}_1$ vertices, where the optimization variables are
the possible $\mathfrak{n}_1$-vertex induced
subgraphs of $G_2$ as well as the possible bijections
between $G_1$ and the induced subgraph,
also restricting that $\phi$ adheres to the seeding function for the seeds.

This formulation is ideal, in that the padding plays exactly the desired
role.
We therefore adopt this formulation for graph matching in the
current setting where the number of vertices in $G_1$ and $G_2$ are different.
The \texttt{SGM} algorithm applied to
$\widetilde{A} \oplus 0_{(\mathfrak{n}_2-\mathfrak{n}_1)\times
(\mathfrak{n}_2-\mathfrak{n}_1)}$ and $\widetilde{B}$ will be referred to as
the \textit{adopted padding scheme}, whereas
applying \texttt{SGM} algorithm to
$A\oplus 0_{(\mathfrak{n}_2-\mathfrak{n}_1)\times
(\mathfrak{n}_2-\mathfrak{n}_1)}$ and $B$ will be referred
to as the \textit{na\"ive padding scheme}.

\begin{figure}[p!]
\centering
\begin{subfigure}[t]{0.3\textwidth}
    \includegraphics[width=1.00\linewidth, clip=true, trim = 20mm 18mm 20mm 10mm ]{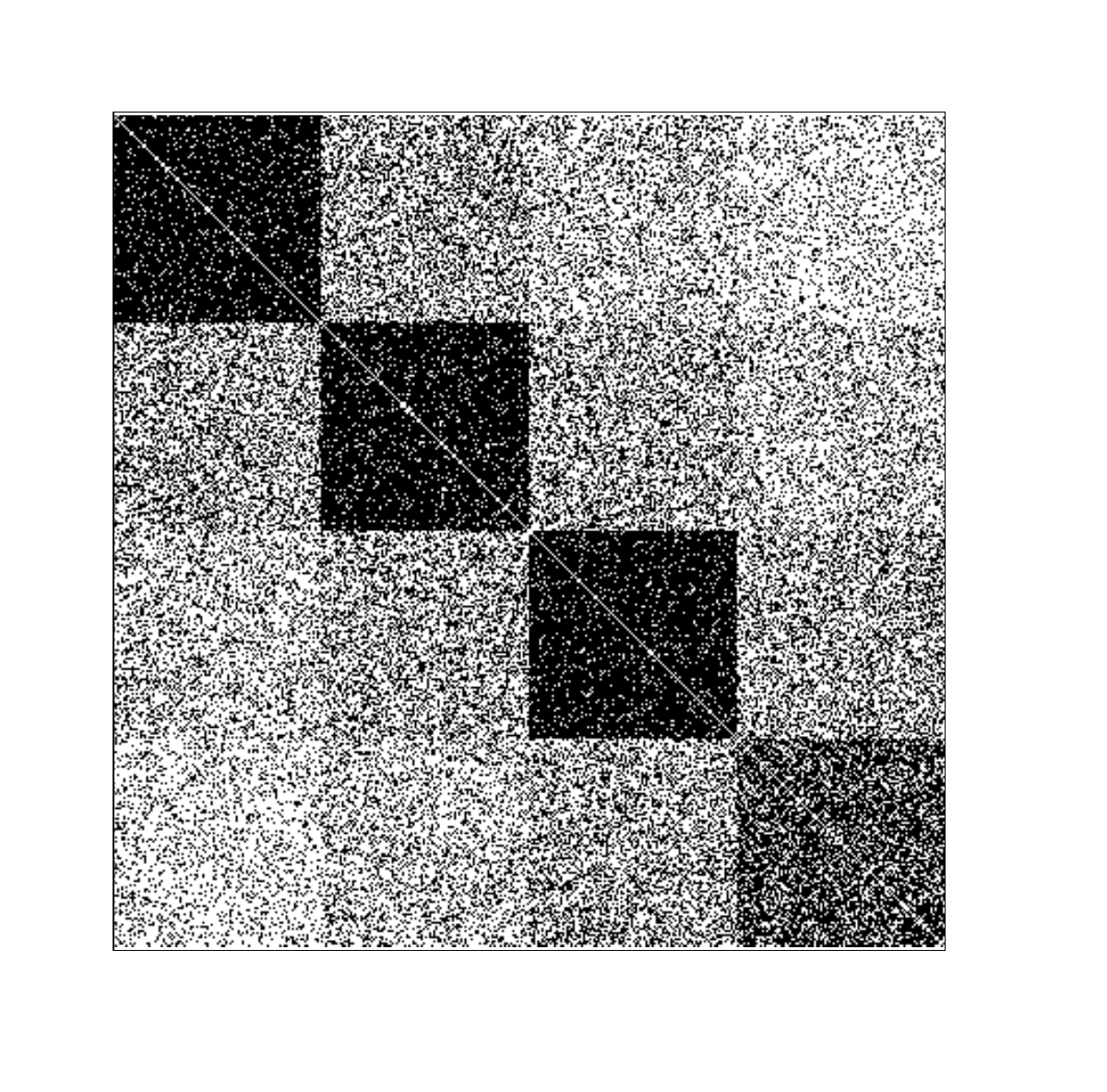}
    \caption{\label{fig:3a}}
\end{subfigure}
\begin{subfigure}[t]{0.3\textwidth}
    \includegraphics[width=1.00\linewidth, clip=true, trim = 20mm 18mm 20mm 10mm]{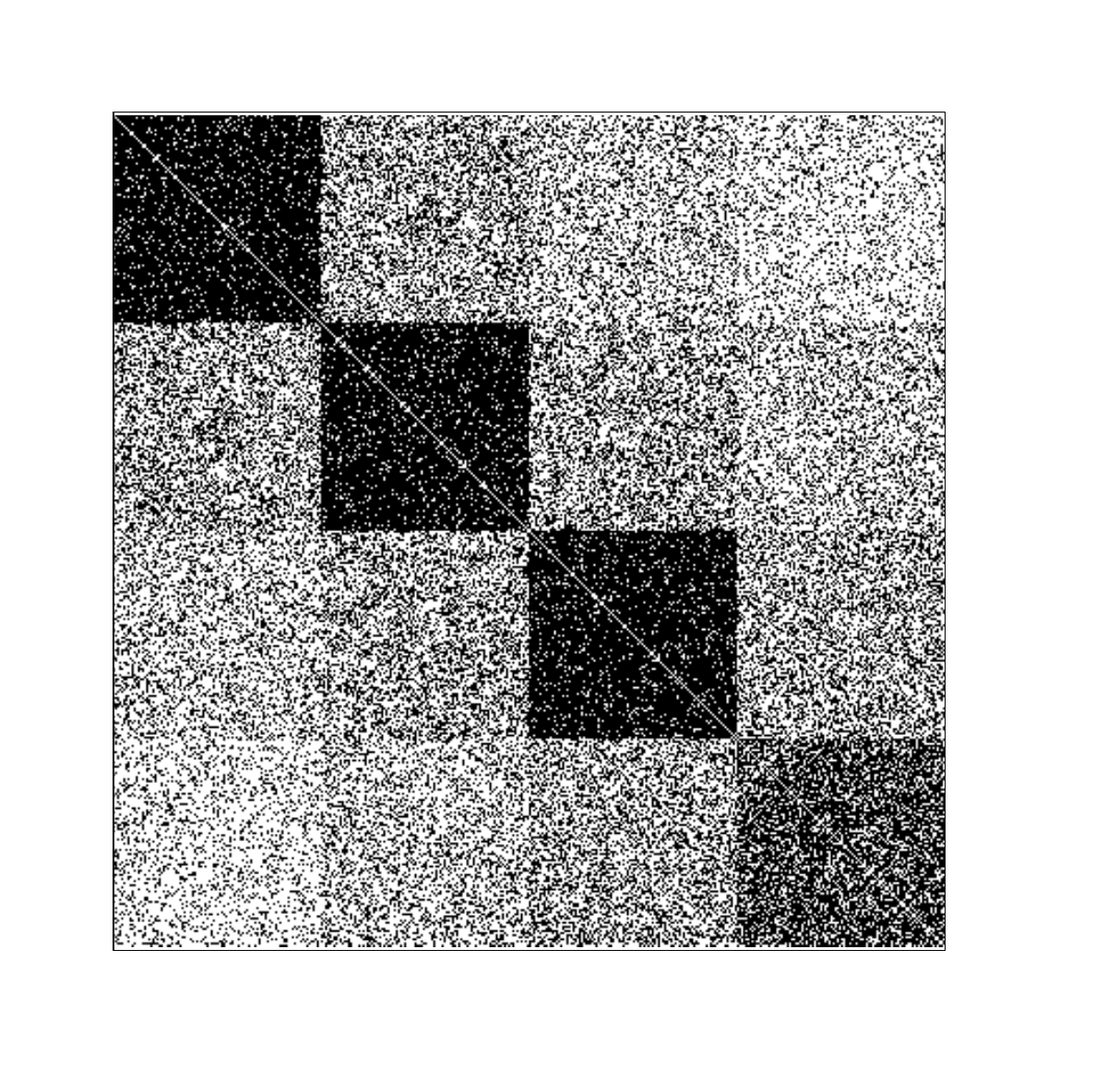}
    \caption{\label{fig:3b}}
\end{subfigure}
\begin{subfigure}[t]{0.3\textwidth}
    \includegraphics[width=1.00\linewidth, clip=true, trim = 20mm 18mm 20mm 10mm]{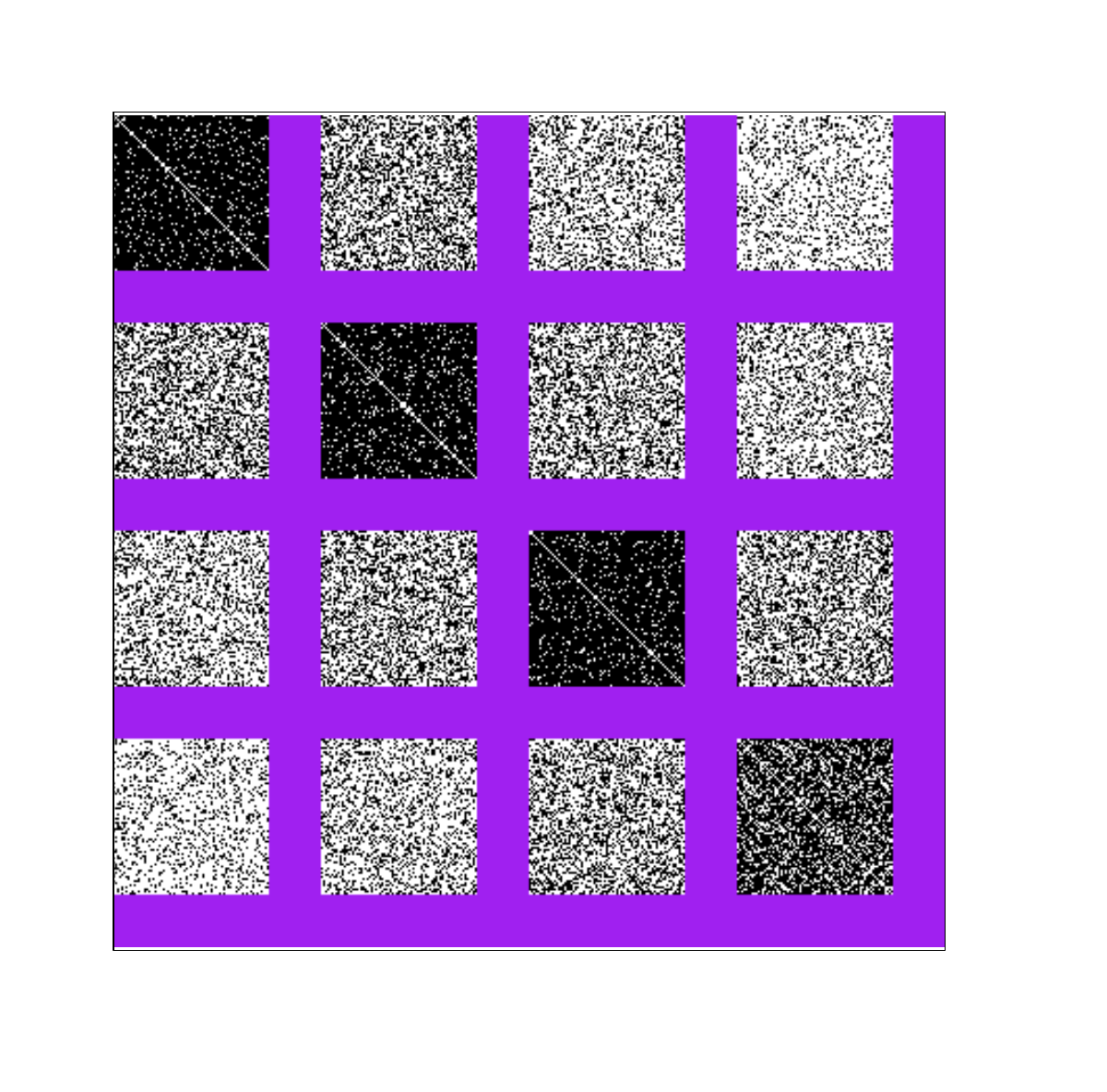}
    \caption{\label{fig:3c}}
\end{subfigure}
\begin{subfigure}[t]{0.3\textwidth}
    \includegraphics[width=1.00\linewidth, clip=true, trim = 20mm 18mm 20mm 10mm]{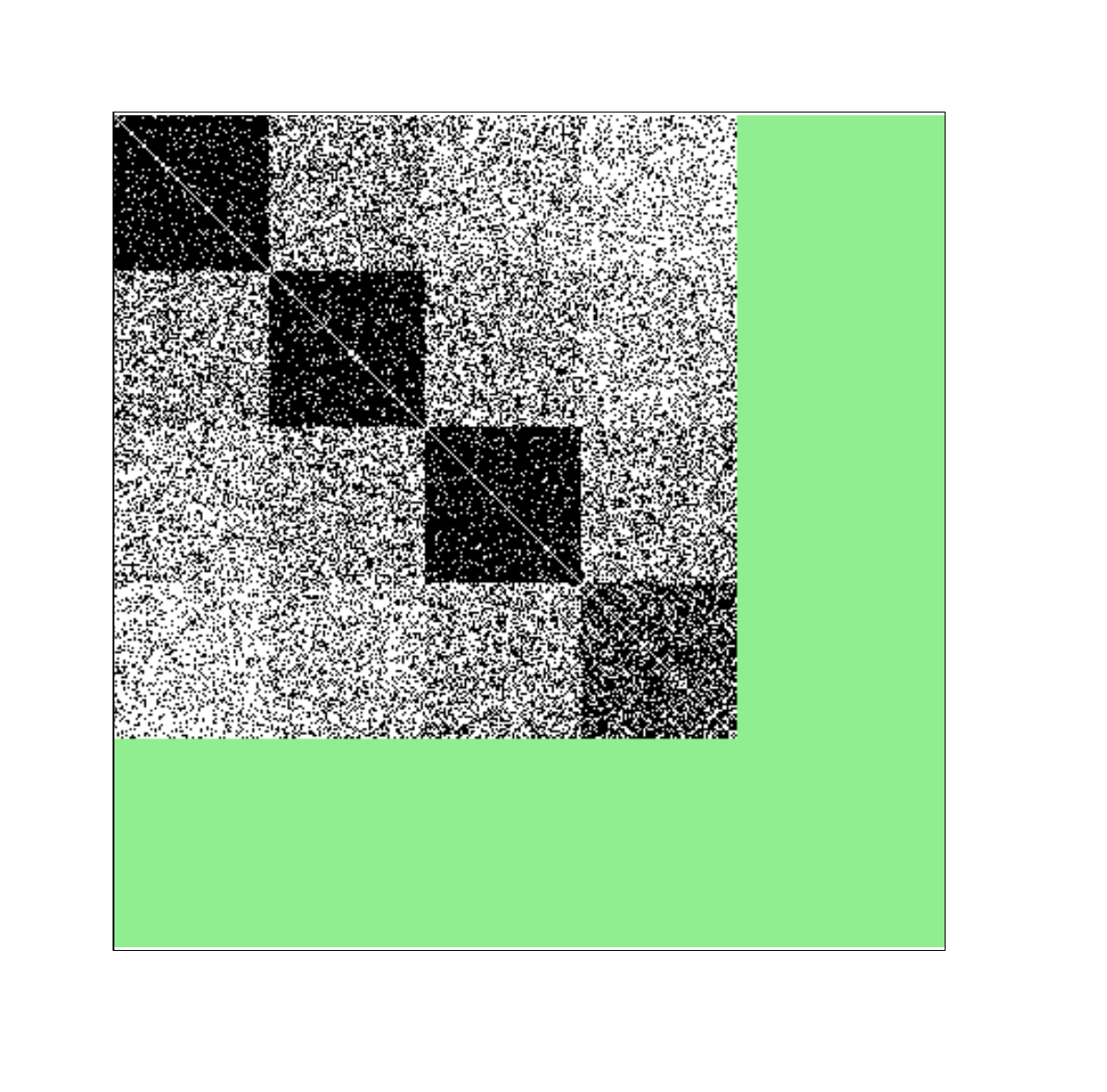}
    \caption{\label{fig:3d}}
\end{subfigure}
\begin{subfigure}[t]{0.3\textwidth}
    \includegraphics[width=1.00\linewidth, clip=true, trim = 20mm 18mm 20mm 10mm]{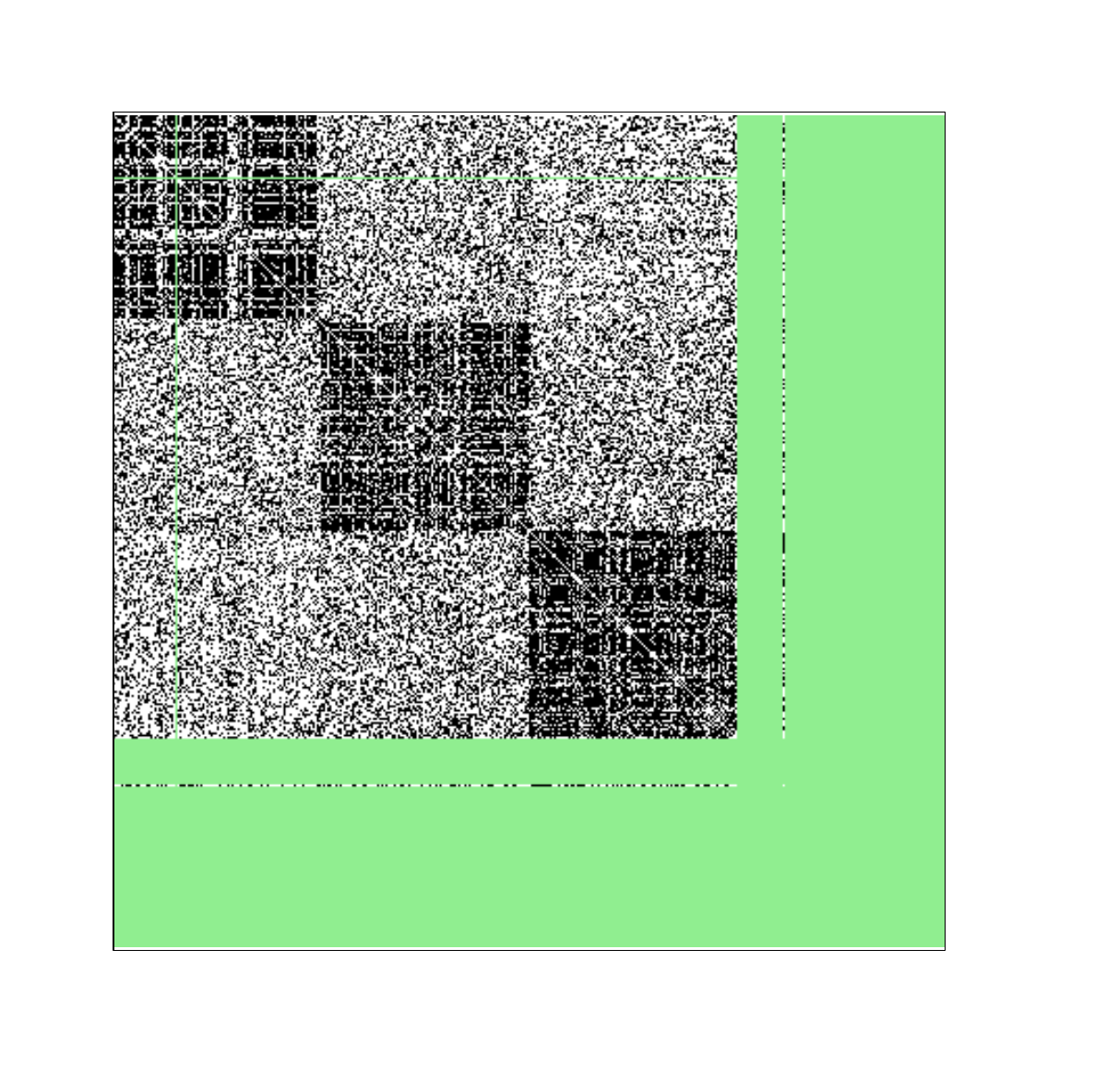}
    \caption{\label{fig:3e}}
\end{subfigure}
\begin{subfigure}[t]{0.3\textwidth}
    \includegraphics[width=1.00\linewidth, clip=true, trim = 20mm 18mm 20mm 10mm]{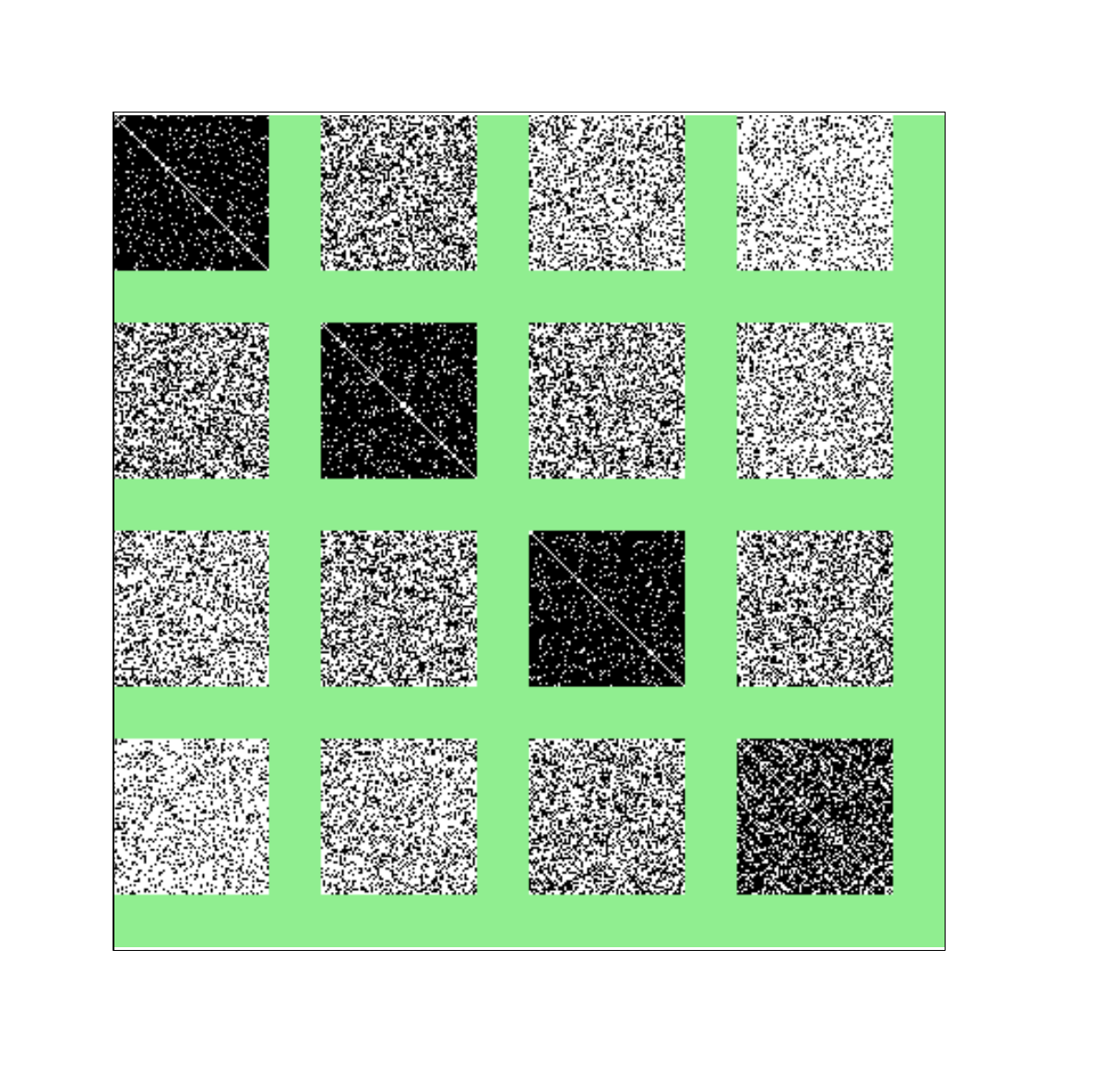}
    \caption{\label{fig:3f}}
\end{subfigure}
\captionsetup{singlelinecheck=off}
\caption{Section \ref{sec:SGMdiff} illustration of na\"ive padding vs adopted padding:\\
{\bf a)} Adjacency matrix for $G_1'$; adjacency in black,
nonadjacency in white. Four blocks of $100$ vertices each,
$400$ vertices total, and the fourth block is less dense
than the other three blocks.\\
{\bf b)} Adjacency matrix for $G_2$; adjacency in black, nonadjacency
in white. Four blocks of $100$ vertices each, $400$ vertices
total, and the fourth block is less dense than the other
three blocks.\\
{\bf c)} Graph $G_1$ is obtained from $G_1'$ 
by deleting $25$
vertices from each block (deleted vertices' rows and
columns in purple).\\
{\bf d)} Adjacency matrix of $G_1$ (which has $300$ vertices) is
(consolidated to) the upper left corner, then padding is
added ($100$ green rows and columns at bottom and
right).\\
{\bf e)} Na\"{i}ve padding; black, white, green in all these
figures have respective values $1,0,0$. Pictured here is
the adjacency matrix of padded-$G_1$, permuted by
\texttt{SGM} to match $G_2$. Padded vertices were
incorrectly paired to the fourth block in $G_2$, which
has $75$ core vertices.\\
{\bf f)} Adopted padding; black, white, green in all these
  figures have respective values $1,-1,0$. Pictured here
  is the adjacency matrix of padded-$G_1$, permuted by
  \texttt{SGM} to match $G_2$. \texttt{SGM} recovered the
  correct correspondences; padding was matched to
  the extraneous vertices.
}
\label{fig:sgmpad}
\end{figure}

To pictorially demonstrate the stark difference between these
two padding schemes, we realized two graphs,
$G_1'$ and $G_2$, each having $400$ vertices,
from a $0.5$-$SBM(4,b,\Lambda)$, where $b$ assigns $100$ vertices
to each of the $k=4$ blocks, and
\[
    \Lambda = \begin{bmatrix}
        .9 & .4 & .3 & .2 \\
        .4 & .9 & .4 & .3 \\
        .3 & .4 & .9 & .4 \\
        .2 & .3 & .4 & .7 \\
    \end{bmatrix}.
\]
Note that the fourth block is less dense than the other three blocks.
The adjacency matrices of $G_1'$ and $G_2$ are respectively pictured in
Figures \ref{fig:3a}, \ref{fig:3b}.
We obtained $G_1$ from $G_1'$ by deleting $25$ uniformly randomly chosen vertices from each block of $G_1'$; 
in Figure \ref{fig:3c} the deleted vertices from $G_1'$ are in purple.
In Figure \ref{fig:3d} the adjacency matrix of $G_1$
is in the upper left corner, and the padding is pictured in green.
After selecting $10$ vertices
uniformly at random from $G_1$ to be seeds, we applied the na\"{i}ve padding scheme 
(wherein black, white, green in these figures have respective values $1,0,0$), as well as the adopted
padding scheme (wherein black, white, green in these figures have
respective values $1,-1,0$). In Figure \ref{fig:3e}, we see that the na\"{i}ve padding scheme maps the padding of $G_1$ to the fourth block of $G_2$, even though $75$ of the vertices
in the fourth block of $G_2$
are core vertices, not extraneous vertices. By contrast, in Figure
\ref{fig:3f} we see that the adopted padding scheme preserves the common
community structure between $G_1$ and $G_2$, and the padding of $G_1$ is mapped to the extraneous vertices of $G_2$.
Indeed, henceforth, we proceed to use the adopted padding scheme and call it
\texttt{Padded SGM}; see Algorithm \ref{alg:padding} for pseudocode.\\

\begin{algorithm}[t!]
\begin{algorithmic}
\STATE \textbf{Input}:
Graphs $G_1$ and $G_2$ with respective vertex sets $\{1,\ldots,\frakn_1\}$ and
$\{1,\ldots,\frakn_2\}$
and with respective adjacency matrices $A$ and $B$,
assuming vertices $\{1,2,\ldots,m\}$ are seeds and assuming $\frakn_1<\frakn_2$

\STATE \textbf{Step 1}:
Compute $\tilde{A} := 2A - J_{\frakn_1}$ and $\tilde{B} :=  2B - J_{\frakn_2}$;

\STATE \textbf{Step 2}:
Apply \texttt{SGM} Algorithm \ref{alg:sgm} to $\tilde{A} \oplus
0_{(\frakn_2-\frakn_1)\times(\frakn_2-\frakn_1)}$ and
$\tilde{B}$ to obtain $\phi_{\hat{P}}$;

\STATE \textbf{return} $\hat{\phi}_r$, the restriction of $\phi_{\hat{P}}$
to the vertices of $G_1$.
\end{algorithmic}
\caption{\texttt{Padded SGM}}
\label{alg:padding}
\end{algorithm}

\vspace{-0.10in}

\begin{figure}[hb!]
\centering
\includegraphics[scale=.3]{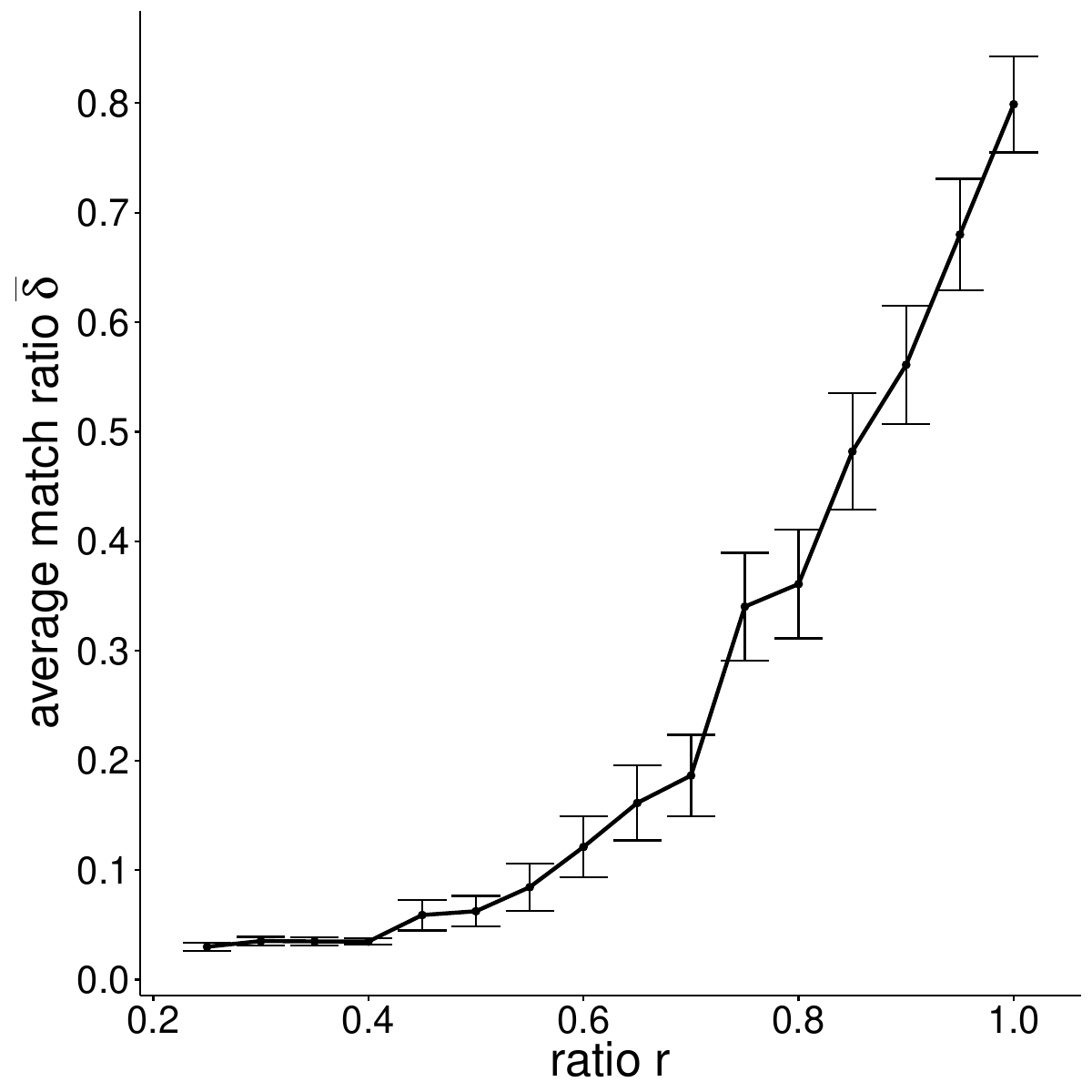}
\vspace*{-0.15in}
\caption{
The average match ratio $\overline{\delta}$ plotted against
the ratio $r=|V_1|/|V_2|=\frakn_1/300$ for the simulation experiment in
Section \ref{sec:SGMdiff}.
\label{fig:ratiovary}}
\end{figure}

We next proceed to utilize Algorithm \ref{alg:padding} \texttt{Padded
SGM} when matching graphs on differently sized vertex sets.
For each value of $r = 0.25,0.30,\ldots,1$,
we realized $300$ pairs of graphs $G_1'$ and $G_2$
from a
$0.7$-SBM($3,b, \Lambda$), where there are $300$ vertices evenly divided
among the three blocks and where $\Lambda$ is given in Equation
(\ref{eqn:lambda}).
$G_1$ is created by deleting vertices in $G_1'$ in such a way
that $G_1$ has $r\cdot 300$ vertices and an equal number of vertices in
each block.
In Figure \ref{fig:ratiovary}, we apply Algorithm \ref{alg:padding}
to $G_1$ and $G_2$ using a randomly selected seed set of size $m=3$, and plot
the mean (with confidence interval)
of the average match ratio $\overline{\delta}$ 
against the ratio $r= \frakn_1 / \frakn_2$.
We note that in the small $r$ settings, \texttt{Padded SGM} poorly recovers the true correspondence between the smaller graph and the core of the larger network.  

While this seems an indictment on our padding approach, we have observed that the matching obtained via Algorithm \ref{alg:padding} can be better \emph{in terms of the objective function being evaluated} than the true permutation in low $r$ settings.
In these cases, graph matching methods are not suited for finding the true correspondence.  Understanding this phenomena further is the subject of future research.

%


\subsection{\texttt{SoftSGM}}
\label{sec:softsgm}

As mentioned, seeded graph matching is an intractable problem.
Thus, because the \texttt{SGM} algorithm is an efficient algorithm,
\texttt{SGM} will not in general find the global
optimal solution; indeed, the realistic goal of \texttt{SGM} is to find a local optimum
that is not too far from the global optimal solution.
Ironically, this shortcoming has a very profitable silver lining,
as we next explain.

Even if we could compute the global optimal solution for seeded graph
matching, there is still a not-insignificant chance that the global
optimal solution is not equal to the natural correspondence function
$\Psi$.
In light of this, it may be quite helpful for a practitioner to have
available several other near-optimal
seeded graph matchings that would provide alternative possibilities, just in
case the practitioner happens to learn through other means that the global optimal match---or a
single highly-touted local optimal match---for
a particular vertex is actually mistaken. It would be even more valuable to be able
to create, for each vertex in one of the graphs, a ranked list of the
vertices in the other graph, ranked by a confidence of being a match.
This can be achieved through sampling local optima near the seeded
graph matching global optimum, and then creating rankings based on the fraction of
time in the sample that pairs of vertices are matched via a local optimum in the sample.

\begin{algorithm}[t!]
\begin{algorithmic}
\STATE \textbf{Input}:
Graphs $G_1, G_2$ each with vertex set $\{1,2,\ldots, m+n\}$, assuming
vertices $\{1,2,\ldots, m\}$ are seeds\\
$R\in\mathbb{N};$ number of restarts of \texttt{SGM} algorithm,\\
$\gamma\in[0,1];$ random initialization step size

\FOR{$i=1:R$}
\STATE \textbf{Step 1}:
Select $Q_i$ uniformly at random from the set of $n\times n$ permutation
matrices;
\STATE \textbf{Step 2}:
Realize $\beta_i$ from a Uniform($0,\gamma$);
\STATE \textbf{Step 3}:
$P^{(0)}_i:=\beta_i Q_i+(1-\beta_i)\frac{1}{n}\vec{1}_{n}\vec{1}_n^T;$
\STATE \textbf{Step 4}:
Apply \texttt{SGM} Algorithm \ref{alg:sgm} to $G_1,G_2$ 
initialized at $P^{(0)}_i$ to obtain $\phi_{\hat{P}}$ 
\STATE \textbf{Step 5}:
Set $P_i := \hat{P}$;
\ENDFOR

\STATE \textbf{Step 6}:
Set $T := I_m \oplus \left( \frac{1}{R} \sum_{i=1}^R P_i\right)$.
\STATE \textbf{return} $T$
\end{algorithmic}
\caption{\texttt{SoftSGM}}
\label{alg:softsgm}
\end{algorithm}

Specifically, in our setting of Section \ref{sec:notation},
given graphs $G_1$ and $G_2$ each with vertex set
$V=\{ 1,2,\ldots,\mathfrak{n}\}$, the
{\it Soft Match SGM} Algorithm (or \texttt{SoftSGM})
consists of running \texttt{SGM} repeatedly on $G_1$,$G_2$ from
randomly sampled starting
doubly stochastic matrices $P^{(0)} \in {\mathcal D}_n$ (where $n$ is
the number of non-seeds), with $R$ denoting the number of such ``restarts."  
Indeed, precisely because \texttt{SGM} will not, in general, solve
the seeded graph matching problem to global optimality,
we will typically obtain a
sample of seeded graph matching approximate solutions, many of which
being different from each other.
For each $i,j \in V$, we will define $T_{ij}$ to be the number
of sample instances in which vertex $i$ in $G_1$ was
matched to vertex $j$ in $G_2$, divided by ($R$) the total number of
sample instances; this fraction will serve as a
confidence for the pair $i,j$ being a match, and these values
create, for each vertex in $G_1$, a probability distribution over possible matches.
For each vertex in $G_1$, ordering the vertices of $G_2$ by decreasing match probability then creates a ranked list of possible matches to the vertices of $G_2$.  
See Algorithm \ref{alg:softsgm} for pseudocode of the \texttt{SoftSGM}
algorithm.

\begin{figure}[t!]
\centering
\begin{subfigure}[b]{0.3\textwidth}
    \includegraphics[width=1.20\linewidth]{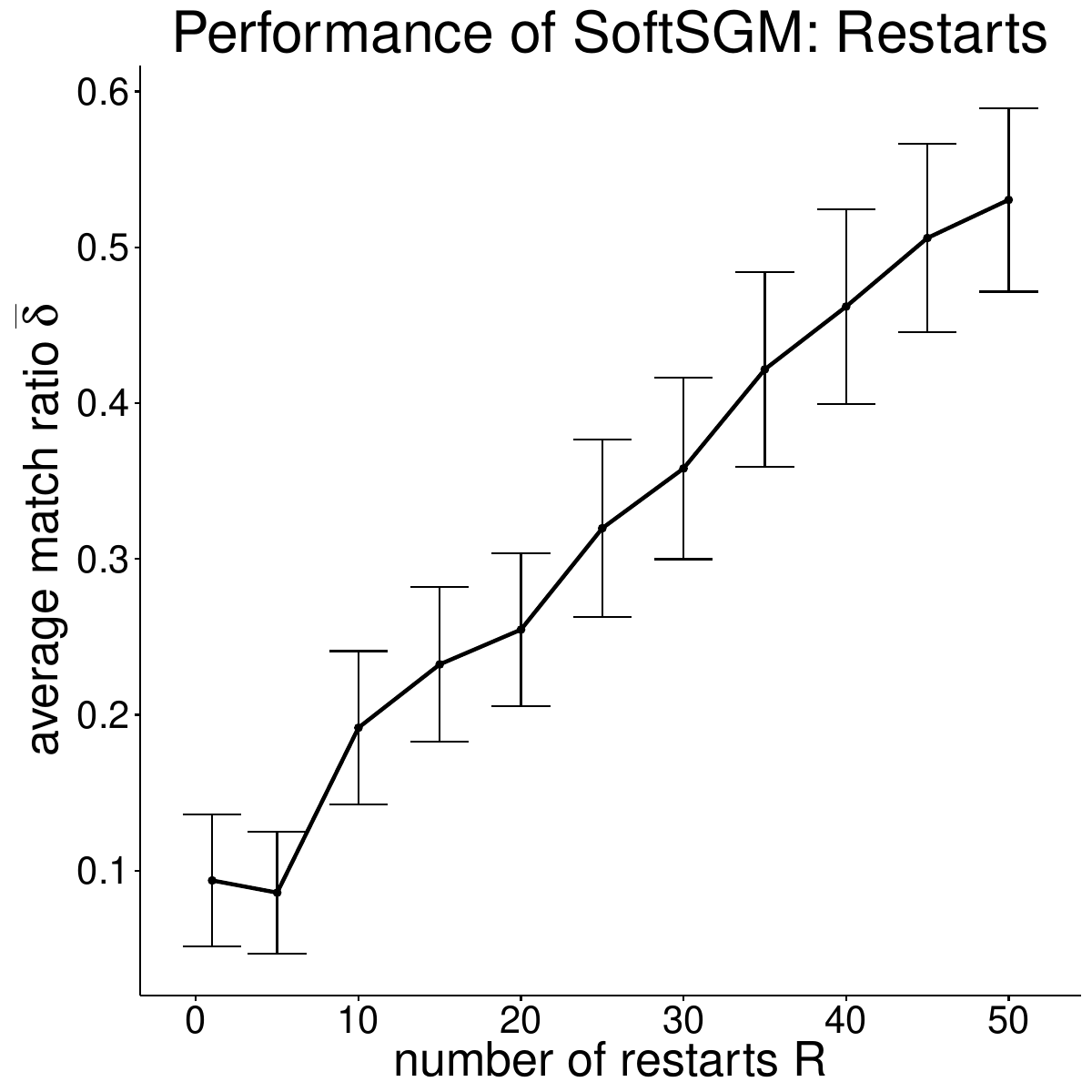}
    \caption{
        $m=3$~seeds, depth~$K=1$, correlation~$\rho=0.6$, varying number
        of restarts~$R$
    \label{fig:restart}
    }
\end{subfigure}
\hspace{15mm}
\begin{subfigure}[b]{0.34\textwidth}
    \includegraphics[width=1.35\linewidth]{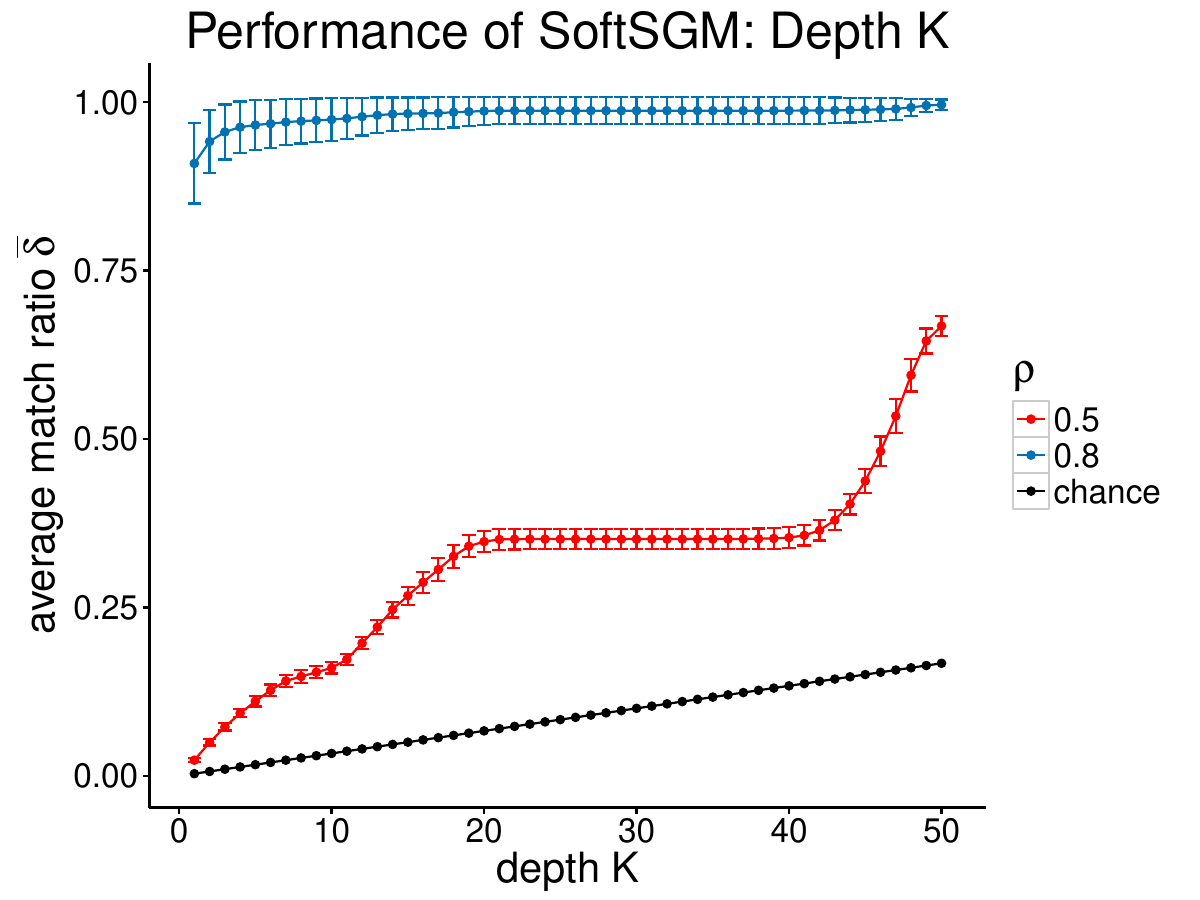}
    \caption{
        $m=0$~seeds, number of restarts $R=50$, varying depth~$K$,
        correlation~$\rho=0.5$~(red), $\rho=0.8$~(blue)
    \label{fig:resCorr}
    }
\end{subfigure}\\
\vspace{-10pt}
    \caption{
        \texttt{SoftSGM} simulations; effects of number of
        restarts $R$ and depth $K$ on average match ratio
        $\overline{\delta}$.
    }
\end{figure}

As a first experiment, we independently realized $50$ pairs of graphs
$G_1$,$G_2$ from a $.7$-SBM($3,b,\Lambda$) distribution with
$\frakn = 30$ vertices,
block assignment function $b$ such that each
of the $k=3$ blocks has $10$ vertices,
$m=4$ seeds discrete-uniformly randomly
selected from the $30$ vertices, and
$\Lambda$ as defined in Equation (\ref{eqn:lambda}). As mentioned before,
for such small graphs we can compute the
global optimum for seeded graph
matching by expressing it as a linear integer
programming problem, and then
solving it exactly with the very powerful software
package {\it GUROBI  Optimizer}.
Among all of the $50 \cdot 30 = 1500$ vertices involved, all but two were
matched (through global optimum from GUROBI) to their correct corresponding vertex.
However, even the two remaining vertices that were not correctly matched
had their corresponding vertex appear in second place in their respective ranked lists
provided by \texttt{SoftSGM} with $R=50$ restarts. In summary, even optimal graph matching might not match a
vertex to its corresponding vertex, and it is useful for a practitioner to have
the ranked list provided by \texttt{SoftSGM} as a recourse.
In the next experiments, the sizes of the graphs are larger, and 
computing the global minimum seeded graph match solution is not 
remotely practical.  

Since \texttt{SoftSGM} provides each vertex with a ranked list of
possible matches (rather than a single proposed matched vertex),
we need to tweak the definition of ``match ratio'' when we want to
measure the effectiveness of \texttt{SoftSGM}.
Let $K \in \mathbb{N}$ be fixed; $K$ will be called the
\textit{match ratio depth}.
In a single \texttt{SoftSGM} simulation, each vertex $i$ in $G_1$ is
called {\it successfully matched at depth $K$} if its corresponding vertex $\Psi(i)$ is
one of the top $K$ vertices in the ranked list associated with $i$.
The match ratio $\delta$ for the simulation is defined to be the
fraction of the non-seed vertices of $G_1$ which are successfully
matched at depth $K$. If multiple simulations are done, then the average match ratio
$\overline{\delta}$ is defined to be the average of the match ratios
over all of the simulations.

To demonstrate how the number of restarts, $R$, impacts the
\texttt{SoftSGM} algorithm,
for each $R=1,5,10,\ldots, 50$,
we independently realize $100$ pairs of graphs from
$0.6$-SBM($3,b,\Lambda$) where
there are $\frakn = 300$ vertices divided evenly into $3$ blocks, with $\Lambda$ given
in Equation (\ref{eqn:lambda}).
We then apply Algorithm \ref{alg:softsgm}, \texttt{SoftSGM},
to each pair of graphs using
$m=3$ uniformly randomly selected seeds, and using depth $K=1$.
We plot the average match ratio $\overline{\delta}$
against the number of restarts $R$ in Figure \ref{fig:restart}.
Note that increasing $R$ yields
dramatic improvement. Of course, when $R=1$ the
\texttt{SoftSGM} algorithm is just implementing Algorithm
\ref{alg:sgm}, since $K=1$ here.

Next, for correlation values $\rho=0.5$
and $\rho=0.8$ 
we apply the \texttt{SoftSGM} algorithm with $R=50$ restarts and no seeds
to $50$ pairs of graphs independently realized
from $\rho-SBM(3,b,\Lambda)$ where the $\frakn = 300$ vertices are evenly
divided into $3$ blocks and $\Lambda$ is as given in Equation
(\ref{eqn:lambda}).
In Figure \ref{fig:resCorr} we plot, for each of $\rho=0.5$ (red curve)
and $\rho=0.8$ (blue curve), the average match ratio
$\overline{\delta}$ as a function of the
depth $K$.
%
This experiment points to the utility of \texttt{SoftSGM} in the limited
seeding regime:
while \texttt{SGM} alone may not recover the true correspondences,
\texttt{SoftSGM} can greatly increase precision at even modest depth,
allowing practitioners the chance to identify the correct correspondences
without having to search the entire vertex set.

\section{Real-Data Demonstrations}
\label{demos}

Thus far we have seen the applicability of the \texttt{SGM} algorithm in
simulated examples involving graphs realized from a $\rho$-correlated
stochastic block model. We now illustrate performance of the \texttt{SGM} algorithm on
three real-data network pairs.\footnote{
The adjacency matrices for our three real-data experiments
are available at \url{http://www.cis.jhu.edu/~parky/SGM}.
}

\subsection{Wikipedia}
\label{wiki}

Wikipedia is an online editable encyclopedia with
(as of April 2017)
40 million articles (more than 5.3 million articles in English) in 293 languages.
A collection of $\frakn=1382$ English articles were collected\footnote{This data set was collected by Dr.\ David J.\
    Marchette in 2009.} by
crawling the (directed) 2-neighborhood of the document ``Algebraic Geometry''
using hyperlinks to traverse from one English article to another.
The vertices in this network are the webpages, with an edge adjoining
one vertex to the next if the first webpage contains a hyperlink to the second.
This first graph was made into a simple undirected
graph by symmetrizing its adjacency matrix.
In Wikipedia, inter-language links between articles of the same topic in different languages are  available;
thus, there is a one-to-one correspondence between the vertices of this
English Wikipedia subgraph and associated vertices of the French Wikipedia graph.
Corresponding articles in French were collected and their intra-language hyperlink structure
yielded a second graph (not necessarily connected) which was also
symmetrized.
The English Wikipedia subgraph is denoted $G_2$ and
the French Wikipedia subgraph induced by the correspondents of the
English Wikipedia articles is denoted $G_1$; thus $G_1$ and $G_2$ both
have 1382 vertices representing webpages, and each webpage in the
English graph has a naturally corresponding webpage in the French graph.

\begin{figure}[ht]
\centering
\vspace*{-1.00in}
\includegraphics[scale=0.35]{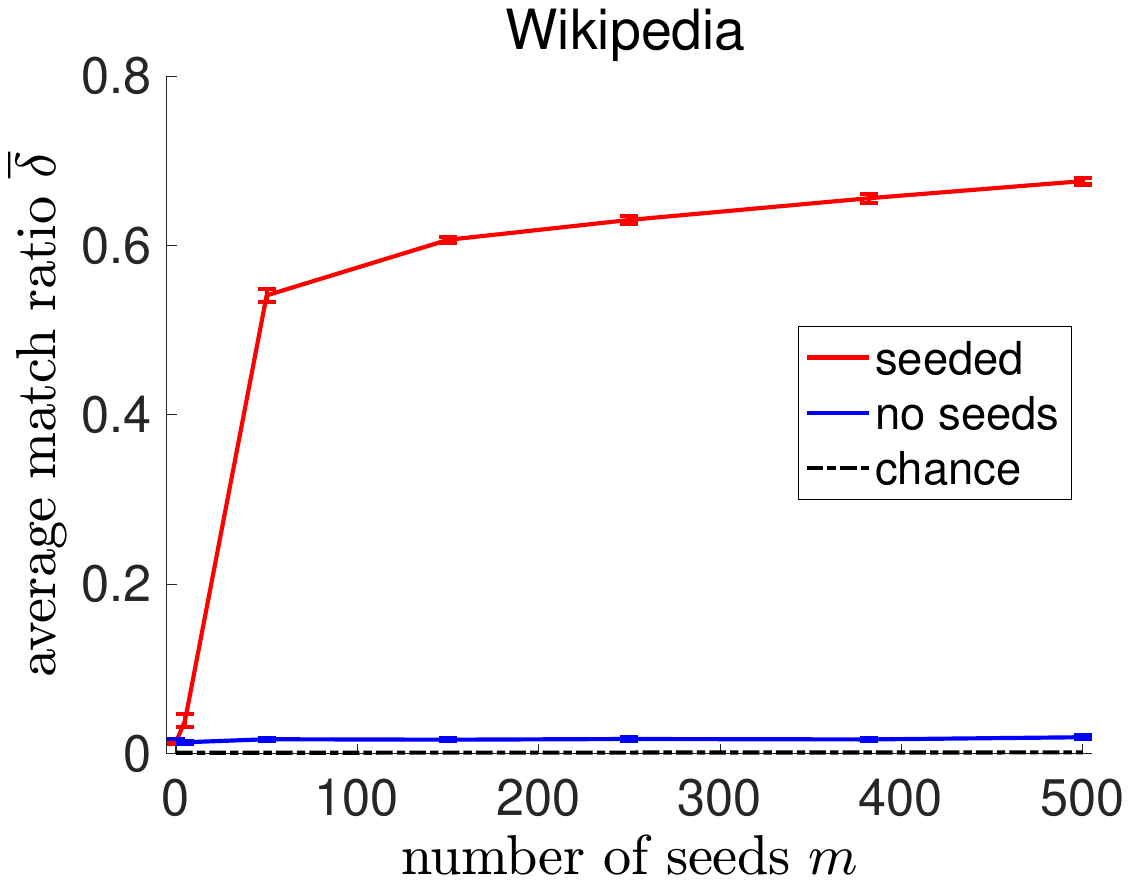}
\vspace*{-1.00in}
\caption{
    Average match ratio $\overline{\delta}$ as a function of number of
    seeds $m$ for the the Wikipedia graph; \texttt{SGM} is red, \texttt{SGM} with no seeds (seeds
    deleted) is blue, chance is black.
\label{Wiki-fig-1}}
\end{figure}

We perform $100$ independent replicates of the following.
For each value of $m = 0,5,50,150,250,382,500$,
we discrete-uniformly randomly select $m$ seeds
and use these seeds for seeded graph matching of the French and English Wikipedia
subgraphs $G_1$ and $G_2$ using the \texttt{SGM} algorithm.
Figure \ref{Wiki-fig-1} depicts, in red, the average match ratio
$\overline{\delta}$ (along with one standard error)
as a function of the number of seeds $m$ and, in black, the
expected average match ratio of chance
(i.e. if vertices were paired uniformly at random), which is
$1/(1382 - m)$.
We see dramatic performance improvement from incorporating just a few seeds:
with no seeds $\overline{\delta} \approx 1/100$ (chance is $1/1382$),
while with just $m=50$ seeds $\overline{\delta} > 1/2$ (chance is $1/1332$).
The blue curve in
Figure \ref{Wiki-fig-1} shows
the average match ratio $\overline{\delta}$
for the {\it unseeded} problem on $\frakn-m$ vertices (i.e. with  the
selected $m$ seeds removed from the two graphs and then applying
\texttt{SGM} with no seeds).
While the problem becomes smaller as $m$ increases,
performance does not improve appreciably.
Of note, when using $m=50$ seeds and using the \texttt{Padded SGM} algorithm
to match the English network with the largest connected component of the French
network (which has 1323 vertices), the match ratio was approximately
0.203.

\begin{figure}[ht]
\centering
    \includegraphics[scale=0.28]{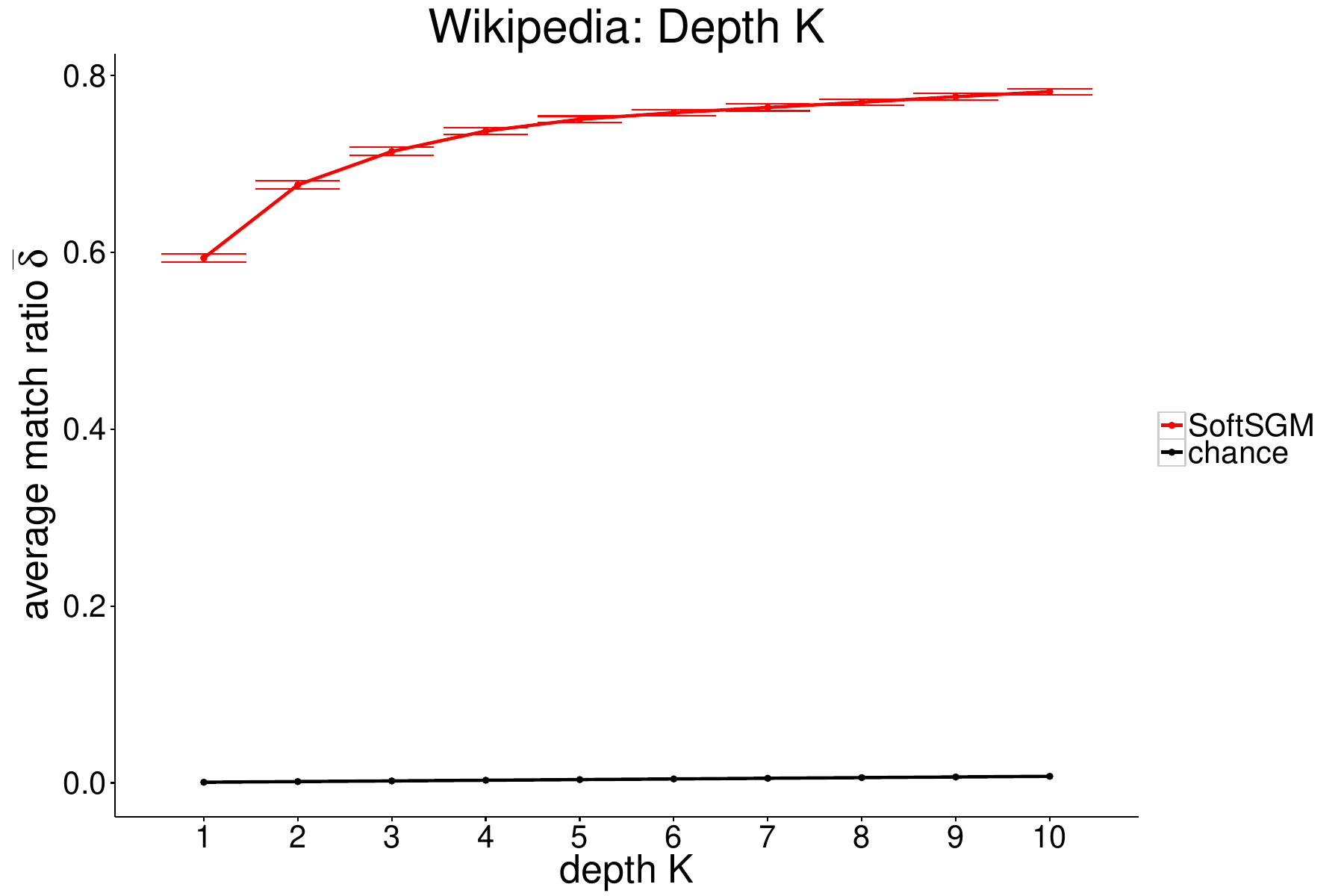}
    \caption{
    \label{fig:wikitopk}
\texttt{SoftSGM} applied to French and English Wikipedia graphs.
Average match ratio $\overline{\delta}$ at depth $K$ plotted using
$m=40$ seeds and $R=25$ restarts. Chance is plotted in black.
    }
\end{figure}

The next experiment consists of discrete-uniformly randomly selecting
$m=40$ seeds in $G_1,G_2$ and applying Algorithm \ref{alg:softsgm}:
\texttt{SoftSGM} using $R=25$ restarts.
We did $25$ independent realizations of this experiment, and Figure
\ref{fig:wikitopk} shows the average match ratio vs depth $K$ for
$K=1,2,\ldots,10$.
Note the improvement brought about by having the ranked list rather
than just a single match for each vertex.
In Figure \ref{fig:wikitopk}, plotted in black is $K/(1382-40)$ for each value
of $K$, which is the expected average match ratio of chance at
depth $K$ (i.e. if the vertices were ordered by uniformly random
permutation).

\subsection{Enron}
\label{enron}

As reported in \cite{enronwiki},
``Enron Corporation was an American energy, commodities, and services
company.
Before its bankruptcy on December 2, 2001,
Enron 
was one of the world's major electricity, natural gas, communications,
and pulp and paper companies, with claimed revenues of nearly \$101
billion during 2000. Fortune named Enron \textit{America's Most Innovative
    Company} for six consecutive years.
At the end of 2001, it was revealed that its reported financial condition was
sustained by institutionalized, systematic, and
creatively planned accounting fraud, known since as the Enron scandal.
Enron has since become a well-known example of willful
corporate fraud and corruption.
The scandal also brought into question the accounting practices and
activities of many corporations in the United States and was a factor in
the enactment of the Sarbanes-Oxley Act of 2002. The scandal also
affected the greater business world by causing the dissolution of the
Arthur Andersen accounting firm.''

In the wake of the Enron Scandal,
the Justice Department released a vast
collection of email messages which have been posted online for academic
use; since privacy constraints usually keep large
collections of email out of reach, this data set is both unique and
valuable to the research community \cite{nytenron}.
The Enron email corpus\footnote{
    See \url{https://en.wikipedia.org/wiki/Enron\_Corpus\#cite\_note-1} for details
and references regarding this data set and other variants.}
consists of messages amongst $\frakn=184$ employees of the Enron Corporation.
Publicly available emails\footnote{
The data we use is available at
\url{http://www.cis.jhu.edu/~parky/SGM}
and \url{http://www.cis.jhu.edu/~parky/Enron/}, and
was obtained from \url{http://www.cs.cmu.edu/~./enron/} in 2004.}
are used to compute a time series of graphs $\{G_t: t=1,\ldots,T\}$
where each graph represents one week of emails in which a node
represents an email address and an edge represents an email sent between
the two addresses during the given week.
An important inference task is to identify ``chatter'' anomalies
--- small groups of actors whose activity amongst themselves increases
significantly for some week $t$ --- since this could potentially indicate
conspiratorial activity amongst the actors.
Previous work has
identified such an anomaly at week $t=132$ (see \cite{PCMP2005}).

\begin{figure}[ht]
\centering
\vspace*{-0.75in}
\includegraphics[scale=0.35]{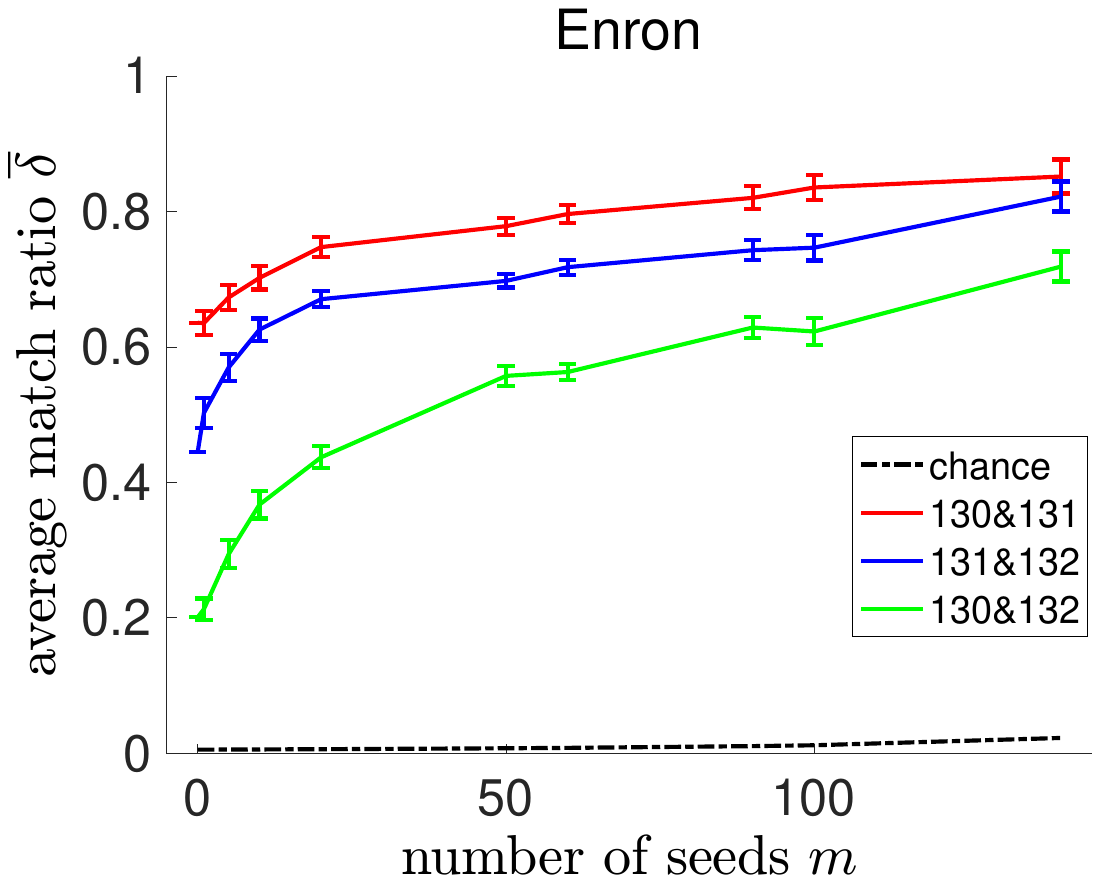}
\vspace*{-0.75in}
\caption{
For each pair of graphs from $\{G_t:t=130,131,132\}$,
the average match ratio $\overline{\delta}$ is plotted against the
number of seeds $m$.
\label{enron-fig}  }
\end{figure}

The Enron email graphs for consecutive weeks $t=130,131,132$
are matched, one pair at a time, using \texttt{SGM}, 
for each of $100$ randomly chosen seed sets of size $m$, for each of
$m = 0, 1, 5, 10, 20, 50, 60,$ $90, 100, 140$.
Figure \ref{enron-fig} plots, for each pair of graphs, the average match
ratio, $\overline{\delta}$, along with one standard error,
against the number of seeds $m$. (Chance is plotted in black.)
The results are consistent with the finding reported in \cite{PCMP2005};
indeed, the average match ratio $\overline{\delta}$ is much higher
between the graphs for weeks $t=130,131$, where there was no significant change,
compared to matching across the change between the graphs for weeks $t=131,132$
and between the graphs for weeks $t=130,132$.
Investigation shows that the difference in performance is largely
attributable to the vertices participating in the anomaly,
as reported in \cite{PCMP2005}.

\begin{figure}[ht]
\centering
    \includegraphics[scale=0.25]{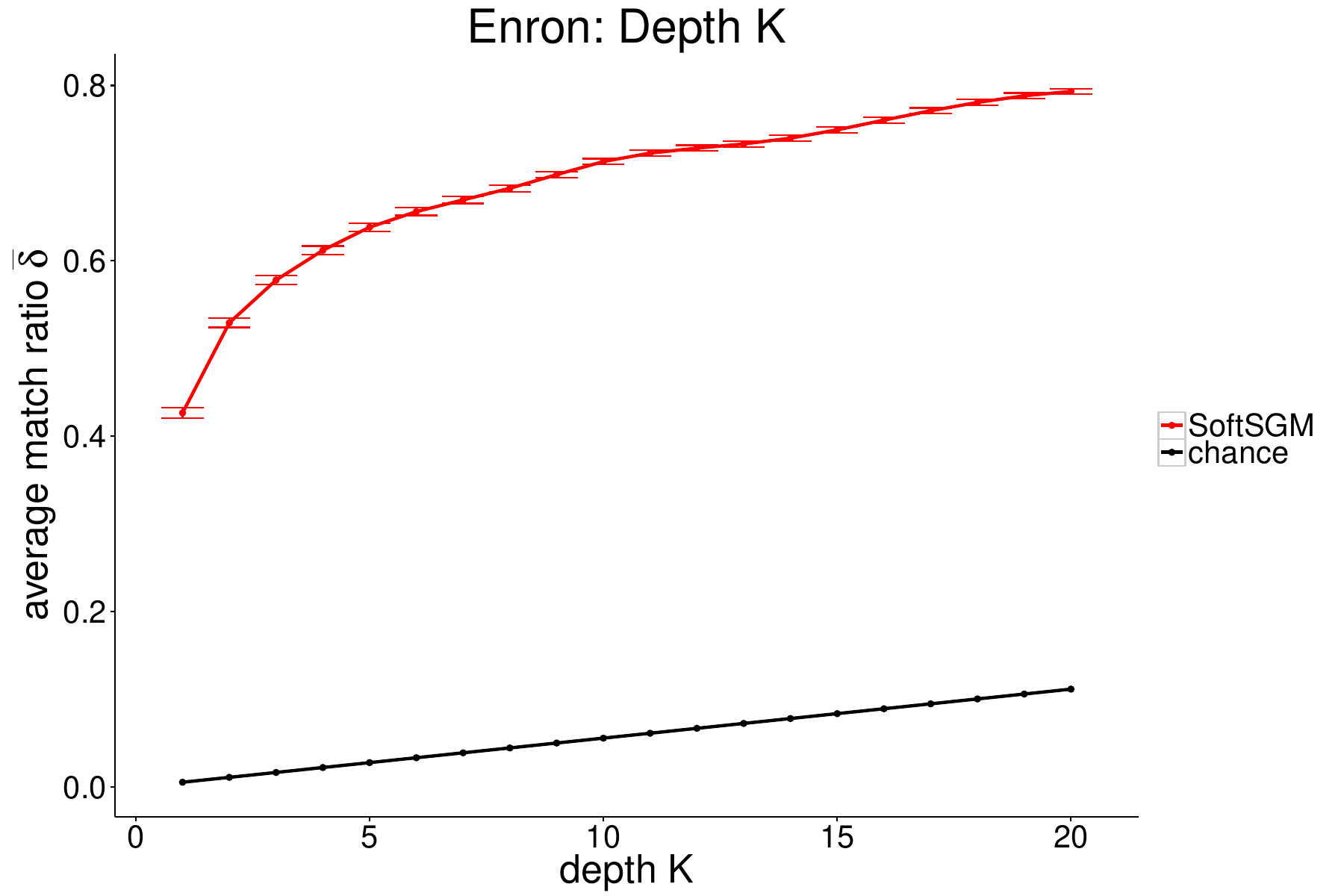}
    \caption{
    \label{fig:enrontopk}
\texttt{SoftSGM} applied to $G_{130}$ and $G_{132}$.
Average match ratio $\overline{\delta}$ at depth $K$ plotted using
$m=5$ seeds and $R=50$ restarts. Chance is plotted in black.
    }
\end{figure}

Next, we do $250$ independent replicates of the following experiment.
\texttt{SoftSGM} is performed on $G_{130},G_{132}$ with $m=5$
discrete-uniformly randomly chosen seeds and $R=50$ restarts. Figure
\ref{fig:enrontopk} plots the average match ratio $\overline{\delta}$
against depth $K=1,2,\ldots,20$.
Note that \texttt{SGM} --- without restarts --- had average match ratio
$\overline{\delta} \approx 0.65$ when using $m=140$ seeds,
but with $R=50$ restarts and depth $K=20$ the average match ratio is
$\overline{\delta} \approx 0.80$ with only $m=5$ seeds.

\subsection{C.\ elegans \label{celegans}}

C.\ elegans is a roundworm that has been extensively studied
(see for example
\cite{lall2006genome,singh2008global,valouev2008high,celegans1}).
We consider $\frakn =279$ neurons in its simple nervous system,
and the connections have been fully mapped in \cite{celegans1};
this mapping was a very important milestone in connectomics.
There are two types of connections between neurons:
chemical synapses and electrical synapses.
We denote by $G_1$ and $G_2$ the graphs for which vertices represent neurons
and edges represent electrical synapses and chemical synapses, respectively.
While these graphs are often weighted and directed, for simplicity and
uniformity with our other examples we take $G_1$ and $G_2$ as
unweighted and undirected.

\begin{figure}[ht]
\centering
\vspace*{-0.75in}
\includegraphics[scale=.35]{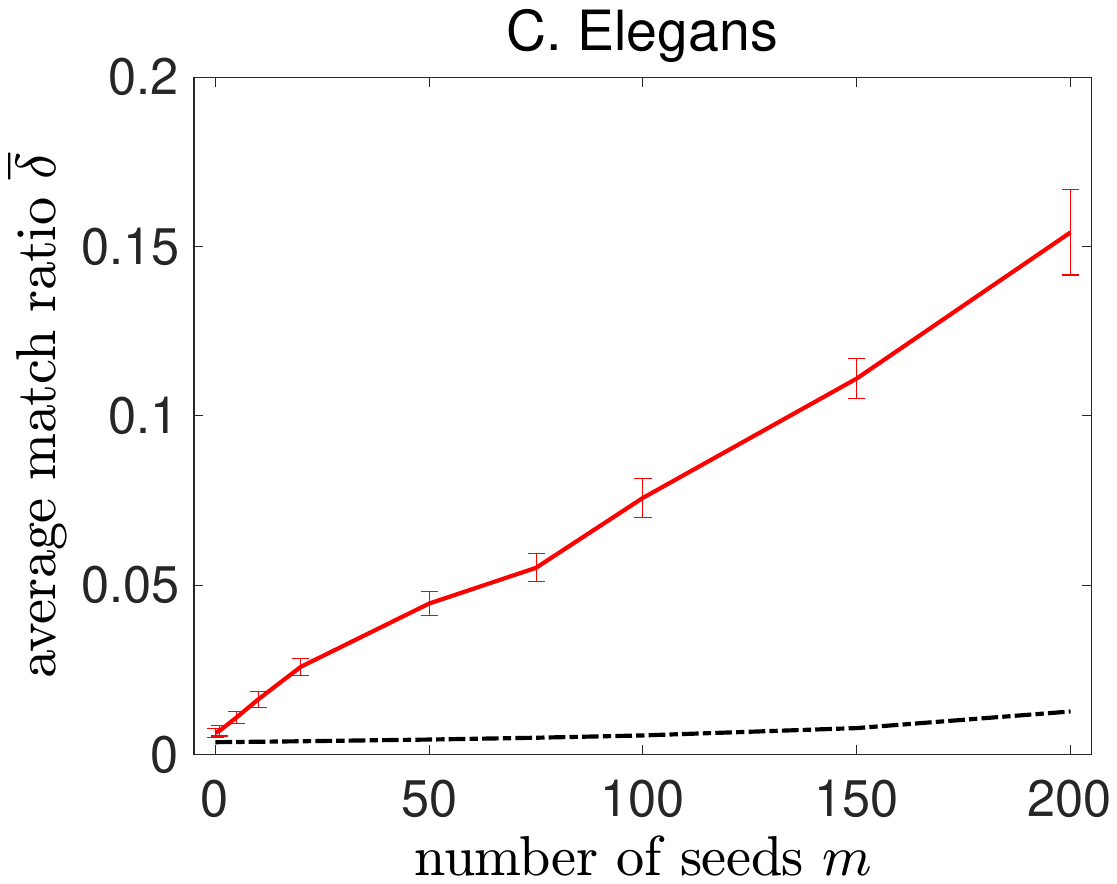}
\vspace*{-0.75in}
\caption{
    The average match ratio $\overline{\delta}$ plotted (in red) against 
    number of seeds $m$ when matching the
    Electrical and Chemical connectivity graphs of C.\ elegans
    nervous system. (Chance is in black.)
\label{worm-fig}}
\end{figure}

For each $m=0,1,5,10,20,50,75,100,150,200$,
we discrete-uniformly randomly select
$100$ sets of $m$ seeds from the vertex set and, for each set of seeds in turn, we
apply Algorithm \ref{alg:sgm}: \texttt{SGM} to the C.\ elegans connectivity
graphs.
In Figure~\ref{worm-fig} we plot the average match ratio 
$\overline{\delta}$ and one standard error (red) against the number of
seeds $m$. (Chance is plotted in black.)
Using \texttt{Padded SGM} to match the
chemical synapse network with the largest connected component of the electrical
synapse network (which has $248$ vertices), the match ratio was approximately
$0.028$ using $m=50$ seeds.

\begin{figure}[ht]
\centering
\begin{subfigure}[b]{0.35\textwidth}
\includegraphics[scale=0.2]{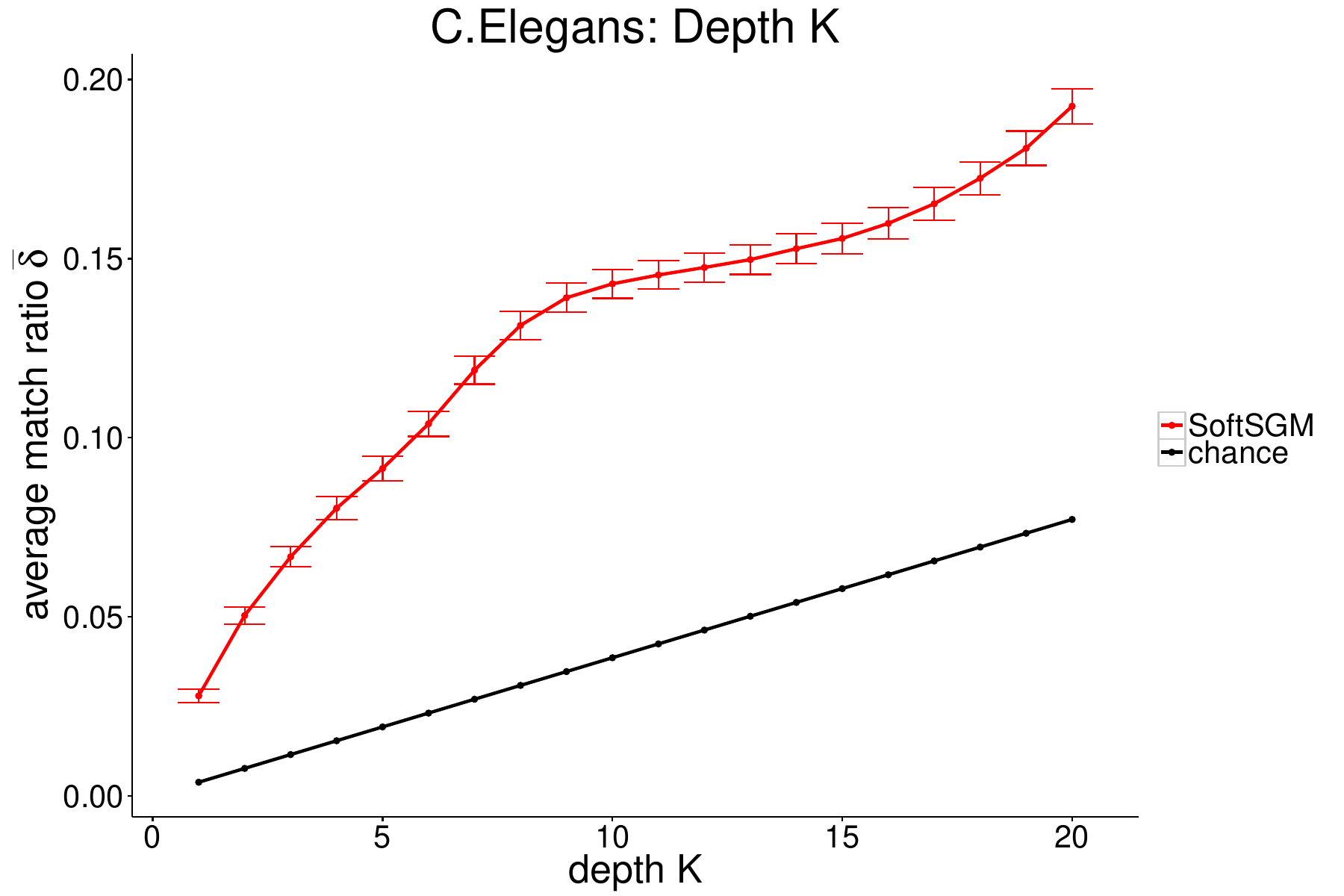}
    \caption{
    \label{fig:celegtopk}
Using Algorithm \ref{alg:softsgm}: \texttt{SoftSGM} with $m=20$ seeds and
$R=50$ restarts for increasing $K$.
    }
\end{subfigure}
\hspace{30mm}
\begin{subfigure}[b]{0.35\textwidth}
    \includegraphics[scale=.2]{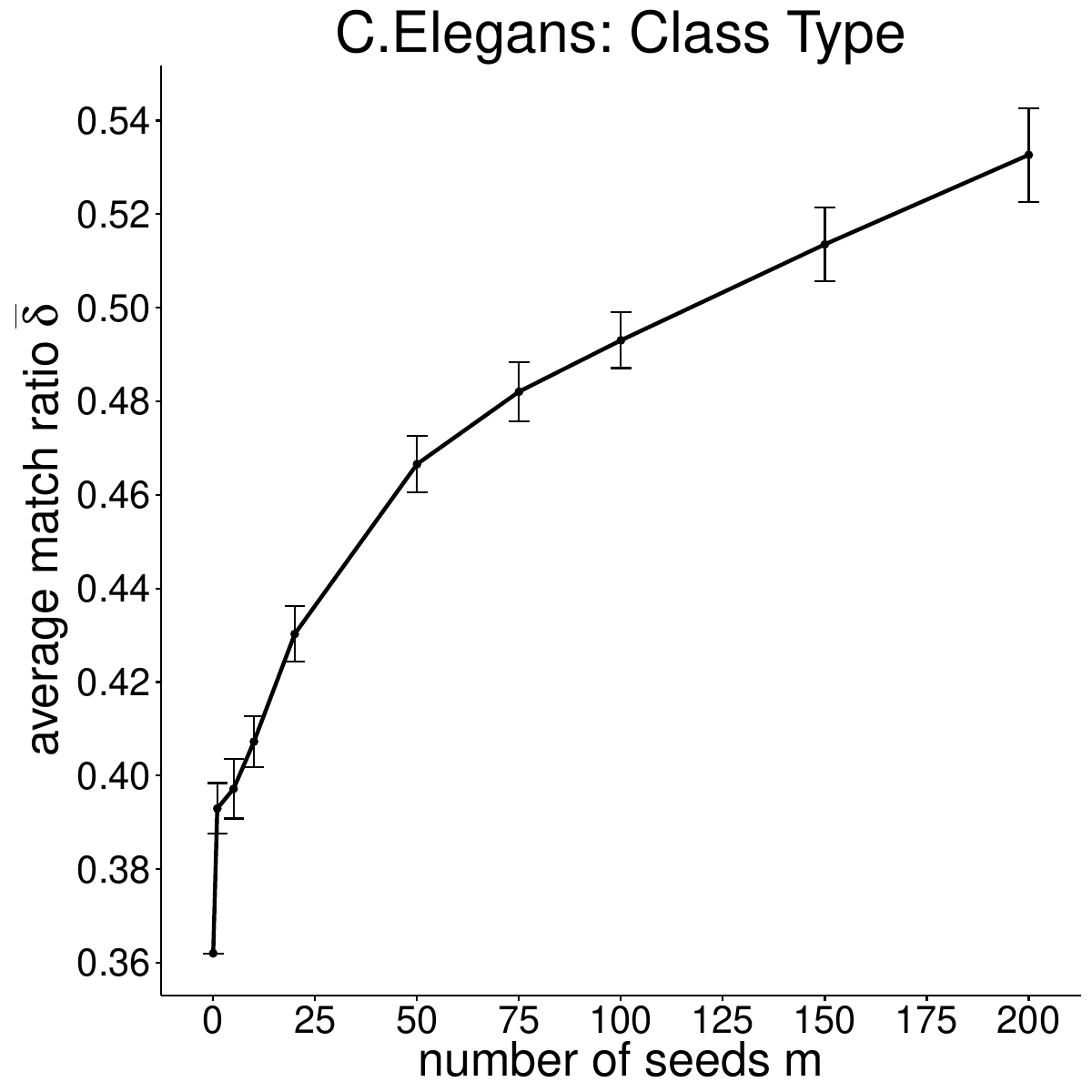}
    \caption{
    \label{fig:celegblock}
Matching classes (inter-, motor, and sensory neuron)
using Algorithm \ref{alg:sgm}: \texttt{SGM} with $m$ seeds.
   }
\end{subfigure}
\caption{
Matching the chemical and electrical networks of
the C.\ elegans worm and plotting performance in terms of the average match
ratio $\overline{\delta}$.
}
\end{figure}

Next, we do $150$ independent replicates of the following experiment.
Algorithm \ref{alg:softsgm}: \texttt{SoftSGM} is used to match
the chemical and electrical networks using $m=20$ seeds and
$R=50$ restarts.
Figure \ref{fig:celegtopk} plots the average match ratio
$\overline{\delta}$ against depth $K=1,2,\ldots,20$.
While only approximately $16\%$ of the vertices can be correctly matched
across the two networks even when using $m=200$ seeds (that is, more
than $2/3$ of the correspondences are already known), nearly $20\%$ of the
vertices can be correctly matched at depth $K=20$ when using only $m=20$
seeds.

Each of the neurons in C. elegans is classified as being either a 
motor neuron, interneuron, or a sensory neuron. For our next
experiment  
we explore what proportion of the C. elegans vertices/neurons  are matched by  
\texttt{SGM} to a vertex/neuron from the same class (ie., a 
motor neuron to a motor neuron, etc.).
For each $m = 0,1,5,10,25,50,75,100,150,200$, we discrete-uniformly 
randomly generate $100$ sets of $m$ seeds and, for each of these seed sets,
we apply \texttt{SGM} to $G_1$, $G_2$. In this context, we consider a vertex 
to be correctly matched if it is matched to a vertex with the same class, and 
average match ratio $\overline{\delta}$ is define accordingly.
In Figure \ref{fig:celegblock} we plot $\overline{\delta}$
as a function of the number of seeds. 
While \texttt{SGM} performs relatively poorly in recovering the exact correspondence between neurons, it performs significantly better in recovering neural classes across networks.
This suggests that the correlation across networks is not at the vertex level, rather at the neuronal class level, with neurons of the same type behaving similarly across network modalities.

\section{Comparing \texttt{SGM} to seeded PATH on QAPLIB}
\label{sec:compare}

As mentioned in Section \ref{intro}, the PATH Algorithm \cite{zaslavskiy2009path} is a popular graph
matching algorithm that is used in practice. The algorithm consists of
considering a sequence of formulations of the underlying problem,
starting with a convex formulation, and ending at a concave formulation,
and the sequence of intermediate formulations traverse a ``path'' along
the convex combination of the initial and final formulation. Approximate
solutions to one formulation serve as the starting point for the next
formulation, and Frank-Wolfe methodology is used on each formulation.

We converted the PATH algorithm to a seeded version, which we call sPATH,
by fixing the alignment variables that correspond to the seeds so that they always
reflect the seed correspondences, and the fixed values of these variables
 are then propagated through the algorithm.

In this section we compare \texttt{SGM} to sPATH on test problems from
QAPLIB, a library of $137$ quadratic assignment problem instances
\cite{burkard1997qaplib}.  Sixteen of these benchmark quadratic
assignment problem instances were deemed particularly challenging, and
were therefore used in \cite{vo_etal15}, \cite{zaslavskiy2009path},
\cite{lyfifivoprsa15} to compare the performance of graph matching
algorithms.  For fifteen of the sixteen we were able to access the
optimal permutation, and not just the optimal objective function value;
of course, for seeding we want this optimal permutation. Note that the
quadratic assignment problem is precisely the optimization problem in
Eq. (\ref{eqn:sgmobj}), except that the maximization is, instead, minimization, and
the matrices $A$ and $B$ are not necessarily adjacency matrices, but can
have any real values as entries. The orders of the matrices in these
fifteen problem instances range from $12$ to $40$.
Of course, modifying the algorithms from maximization to minimization is
trivially accomplished by appropriate negations.

For each of the fifteen benchmark quadratic assignment problem instances, and for
each value of $m=1,2,3,4$, we did $30$ independent experiments
of randomly selecting $m$ seeds, and we compared the average (over the thirty
experiments) objective function values of \texttt{SGM} to sPATH and to the optimal solution,
see Table \ref{table:qap} for these results. On these benchmarks, \texttt{SGM} was more effective than sPATH.

The PATH algorithm code we accessed \cite{zaslavskiy2009path} was written in C++, and we seeded it
directly in its native C++, writing the adjustments into the original code,
whereas our SGM algorithm is written in R. (Also, while SGM was run on a laptop
with Intel Core i5 @ 1.6 GHz, sPATH was run on the MARCC computer with specs as given in subsection \ref{ssec:time}.)
The average runtimes for the experiments 
on the larger benchmarks were
on the order of $0.01$ seconds for SGM and 
$0.14$ seconds for PATH. Because the runtimes
for sPATH are an order of magnitude greater than for SGM (and are indeed so, more generally),
sPATH is not comparable more broadly to SGM, and we therefore do not include more comparison.

\begin{table}[ht]
    \tiny
    \setlength{\arraycolsep}{1.5pt}
\centering
\begin{tabular}{|r|l|r|r|r|r|r|r|r|r|r}
  \hline
 \multirow{2}[-3]*{QAP}  & \multirow{2}[-3]*{OPT}  &
 \multicolumn{2}{|c|}{$m=1$} & \multicolumn{2}{|c|}{$m = 2$} &
 \multicolumn{2}{|c|}{$m = 3$} &\multicolumn{2}{|c|}{$m=4$} 
 \\ 
\hline
& & SGM & sPATH & SGM & sPATH & SGM & sPATH & SGM & sPATH 
\\ 
  \hline
 chr12c & 11156 & 18770 & 31858 & 16298 & 30889 & 16221 &
 27031 & 14504 & 25073 
 \\ 
 chr15a & 9896 & 15813 & 49522 & 16280 & 42959 & 15861 & 35591 & 14399 & 32041 
 \\ 
 chr15c & 9504 & 18230 & 45144 & 15649 & 40329 & 15494 & 36909 & 14027 & 35654 
 \\ 
 chr20b & 2298 & 3555 & 9411 & 3585 & 8991 & 3540 & 8091 & 3556 & 7521 
 \\ 
 chr22b & 6194 & 8359 & 14075 & 8184 & 13503 & 8021 & 13126 & 7673 & 12475 
 \\ 
 esc16b & 292 & 293 & 308 & 295 & 304 & 294 & 301 & 293 & 301 
 \\ 
 rou12  & 235528 & 250799 & 285085 & 242999 & 276292 & 236993 & 268722 & 236432 & 264059 
 \\ 
 rou15  & 354210 & 369198 & 449821 & 361721 & 439908 & 357969 & 415076 & 356537 & 402067 
 \\ 
 rou20  & 725522 & 748128 & 863811 & 753645 & 886876 & 746137 & 855938 & 743474 & 850498 
 \\ 
 tai15a & 388214 & 403314 & 463836 & 402760 & 461114 & 398765 & 449411 & 396544 & 438072 
 \\ 
 tai17a & 491812 & 518678 & 590697 & 506259 & 596524 & 506159 & 575686 & 502410 & 563565 
 \\ 
 tai20a & 703482 & 736797 & 855532 & 739771 & 852671 & 735472 & 850778 & 716565 & 828659 
 \\ 
 tai30a & 1818146 & 1888526 & 2141265 & 1878886 & 2123362 & 1874521 & 2119654 & 1865151 & 2109946 
 \\ 
 tai35a & 2422002 & 2515301 & 2876351 & 2505556 & 2838981 & 2504548 & 2812018 & 2493860 & 2800284 
 \\ 
 tai40a & 3139370 & 3255807 & 3716363 & 3261394 & 3662562 & 3246184 & 3630483 & 3249476 & 3611428 
 \\ 
\hline
\end{tabular}
\caption{Average objective function from minimization via SGM and sPATH (as well as the optimal solution) on
    15 benchmark examples of the QAPLIB library, with $m=1,2,3,4$ seeds.
    \label{table:qap}
}
\end{table}

\section{Discussion}
\label{disc}

Many graph inference tasks involve multiple graphs consisting of
corresponding vertex sets. These tasks are more easily accomplished
if we know the correspondence between the vertices in the different graphs,
but these correspondences might be hidden from us.
Data fusion based on such a correspondence, if it can be (at least
approximately) recovered, will result in data
representation that reveals hidden relationships between graph vertices, providing a more complete worldview.

In this manuscript we modify the
(theoretically principled and computationally tractable)
graph matching \texttt{FAQ} algorithm of Vogelstein et al (2015) 
\cite{vo_etal15}
to obtain what we call the \texttt{SGM} algorithm and its variants, which
\begin{enumerate}
    \item Incorporate the use of seeds,
    \item Can match graphs of different orders (i.e., are differently sized), and
    \item Can provide a soft matching which assigns ---to each pair of vertices across the two
        graphs--- a value representing our confidence that the pair correspond.
\end{enumerate}
We demonstrated the effectiveness of the \texttt{SGM} algorithm and its variants
via simulations and three real data experiments. In particular, seeding can provide a
dramatic increase in the success of recovering underlying vertex correspondence. Also,
soft matching provides each vertex with a ranked list of potential correspondents
instead of a single proposed correspondent, which leaves the practitioner with recourse
in the event of discovery (by other means) that a proposed correspondent is not correct.

In practice, identifying seeds may be costly. Thus, it will be important to understand
the cost-benefit trade-off between inference without correspondence vs inference
performed subsequent to seed discovery and utilization.
This paper provides the foundation for that analysis. Note that the value of a few seeds leads to the
demand for an {\it active learning} methodology to identify the most cost-effective
vertices to use as seeds, for example see \cite{lica15}.

Obvious extensions to this work include:
(a) the case where the correspondence may be many-to-many;
and (b) the case where the seeds themselves are soft/errorful;
this means that we know that it is {\it likely} (but not {\it certain})
that various pairs of vertices correspond.
Each of these extensions can be addressed within the framework presented here.

Also, there are more general notions of closeness between graphs, such as 
graph edit distance. See \cite{bougleux2017graph} for analysis and 
a successful algorithmic approach for this form of graph matching. 
It would be profitable to consider ways to seed this algorithm.

In conclusion, we contend that the methodology presented herein
forms the foundation for improving performance in myriad graph inference applications
for which there exists a partially \emph{known-or-discoverable}
correspondence between the vertices of various graphs.



\bibliographystyle{ieeetr}
\bibliography{sgm12bib}

\begin{thebibliography}{10}

\bibitem{lu2016fast}
Y.~Lu, K.~Huang, and C.-L. Liu, ``A fast projected fixed-point algorithm for
  large graph matching,'' {\em Pattern Recognition}, vol.~60, pp.~971--982,
  2016.

\bibitem{riesen2016predicting}
K.~Riesen and M.~Ferrer, ``Predicting the correctness of node assignments in
  bipartite graph matching,'' {\em Pattern Recognition Letters}, vol.~69,
  pp.~8--14, 2016.

\bibitem{lerouge2017new}
J.~Lerouge, Z.~Abu-Aisheh, R.~Raveaux, P.~H{\'e}roux, and S.~Adam, ``New binary
  linear programming formulation to compute the graph edit distance,'' {\em
  Pattern Recognition}, vol.~72, pp.~254--265, 2017.

\bibitem{sang2012robust}
J.~Sang and C.~Xu, ``Robust face-name graph matching for movie character
  identification,'' {\em IEEE Transactions on Multimedia}, vol.~14, no.~3,
  pp.~586--596, 2012.

\bibitem{egozi2013probabilistic}
A.~Egozi, Y.~Keller, and H.~Guterman, ``A probabilistic approach to spectral
  graph matching,'' {\em IEEE Transactions on Pattern Analysis and Machine
  Intelligence}, vol.~35, no.~1, pp.~18--27, 2013.

\bibitem{iodice2015salient}
S.~Iodice and A.~Petrosino, ``Salient feature based graph matching for person
  re-identification,'' {\em Pattern Recognition}, vol.~48, no.~4,
  pp.~1074--1085, 2015.

\bibitem{pedarsani2011privacy}
P.~Pedarsani and M.~Grossglauser, ``On the privacy of anonymized networks,'' in
  {\em Proceedings of the 17th ACM SIGKDD international conference on Knowledge
  discovery and data mining}, pp.~1235--1243, ACM, 2011.

\bibitem{yartseva2013performance}
L.~Yartseva and M.~Grossglauser, ``On the performance of percolation graph
  matching,'' in {\em Proceedings of the first ACM conference on Online social
  networks}, pp.~119--130, ACM, 2013.

\bibitem{chvolypr16joint}
L.~Chen, J.~T. Vogelstein, V.~Lyzinski, and C.~E. Priebe, ``A joint graph
  inference case study: The {C}.{E}legans chemical and electrical
  connectomes,'' in {\em Worm}, vol.~5, p.~e1142041, Taylor \& Francis, 2016.

\bibitem{yang2009graph}
F.~Yang and F.~Kruggel, ``A graph matching approach for labeling brain sulci
  using location, orientation, and shape,'' {\em Neurocomputing}, vol.~73,
  no.~1-3, pp.~179--190, 2009.

\bibitem{cofosave04}
D.~Conte, P.~Foggia, C.~Sansone, and M.~Vento, ``Thirty years of graph matching
  in pattern recognition,'' {\em International Journal of Pattern Recognition
  and Artificial Intelligence}, vol.~18, no.~3, pp.~265--298, 2004.

\bibitem{fopeve14}
P.~Foggia, G.~Perncannella, and M.~Vento, ``Graph matching and learning in
  pattern recognition in the last 10 years,'' {\em Internation Journal of
  Pattern Recognition and Artificial Intelligence}, vol.~28, no.~1, 2014.

\bibitem{yaetal16}
J.~Yan, X.-C. Yin, W.~Lin, C.~Deng, H.~Zha, and X.~Yang, ``A short survey of
  recent advances in graph matching,'' in {\em Proceedings of the 2016 ACM on
  International Conference on Multimedia Retrieval}, pp.~167--174, ACM, 2016.

\bibitem{sago1976}
S.~Sahni and T.~Gonzalez, ``P-complete approximation problems,'' {\em Journal
  of the ACM (JACM)}, vol.~23, no.~3, pp.~555--565, 1976.

\bibitem{ba15}
L.~Babai, ``Graph isomorphism in quasipolynomial time,'' {\em arXiv preprint
  arXiv:1512.03547}, 2016.

\bibitem{zaslavskiy2009global}
M.~Zaslavskiy, F.~Bach, and J.-P. Vert, ``Global alignment of protein--protein
  interaction networks by graph matching methods,'' {\em Bioinformatics},
  vol.~25, no.~12, pp.~1259--1267, 2009.

\bibitem{type_constraints}
C.~Fraikin and P.~Van~Dooren, ``Graph matching with type constraints on nodes
  and edges,'' in {\em Dagstuhl Seminar Proceedings}, (Dagstuhl, Germany),
  Schloss Dagstuhl-Leibniz-Zentrum f{\"u}r Informatik, 2007.

\bibitem{ham2005semisupervised}
J.~Ham, D.~D. Lee, and L.~K. Saul, ``Semisupervised alignment of manifolds.,''
  in {\em AISTATS}, pp.~120--127, 2005.

\bibitem{lyzinski2015spectral}
V.~Lyzinski, D.~L. Sussman, D.~E. Fishkind, H.~Pao, L.~Chen, J.~T. Vogelstein,
  Y.~Park, and C.~E. Priebe, ``Spectral clustering for divide-and-conquer graph
  matching,'' {\em Parallel Computing}, vol.~47, pp.~70--87, 2015.

\bibitem{lica15}
L.~Li and W.~M. Campbell, ``Matching community structure across online social
  networks,'' {\em Network NIPS}, 2015.

\bibitem{hurugu13}
N.~Hu, R.~M. Rustamov, and L.~Guibas, ``Graph matching with anchor nodes: A
  learning approach,'' pp.~2906--2913, 2013.

\bibitem{kahagr15}
E.~Kazemi, S.~H. Hamed, and M.~Grossglauser, ``Growing a graph matching from a
  handful of seeds,'' {\em Proceedings of the VLDB Endowment}, vol.~8, no.~10,
  pp.~1010--1021, 2015.

\bibitem{vo_etal15}
J.~T. Vogelstein, J.~M. Conroy, V.~Lyzinski, L.~J. Podrazik, S.~G. Kratzer,
  E.~T. Harley, D.~E. Fishkind, R.~J. Vogelstein, and C.~E. Priebe, ``Fast
  approximate quadratic programming for graph matching,'' {\em PLOS one},
  vol.~10, no.~4, p.~e0121002, 2015.

\bibitem{lyadvopapr14}
V.~Lyzinski, S.~Adali, J.~T. Vogelstein, Y.~Park, and C.~E. Priebe, ``Seeded
  graph matching via joint optimization of fidelity and commensurability,''
  {\em arXiv preprint arxiv:1401.3813}, 2014.

\bibitem{lyfipr14}
V.~Lyzinski, D.~Fishkind, and C.~Priebe, ``Seeded graph matching for correlated
  {E}rd{\H{o}}s-{R}{\`e}nyi graphs,'' {\em Journal of Machine Learning
  Research}, vol.~15, pp.~3693--3720, 2014.

\bibitem{lysutaatpr14perfectase}
V.~Lyzinski, D.~L. Sussman, M.~Tang, A.~Athreya, and C.~E. Priebe, ``Perfect
  clustering for stochastic blockmodel graphs via adjacency spectral
  embedding,'' {\em Electronic Journal of Statistics}, vol.~8, no.~2,
  pp.~2905--2922, 2014.

\bibitem{filypachpr15}
D.~E. Fishkind, V.~Lyzinski, H.~Pao, L.~Chen, and C.~E. Priebe, ``Vertex
  nomination schemes for membership prediction,'' {\em The Annals of Applied
  Statistics}, vol.~9, no.~3, pp.~1510--1532, 2015.

\bibitem{lyzinski2016information}
V.~Lyzinski, ``Information recovery in shuffled graphs via graph matching,''
  {\em arXiv preprint arXiv:1605.02315}, 2016.

\bibitem{lyfifivoprsa15}
V.~Lyzinski, D.~E. Fishkind, M.~Fiori, J.~T. Vogelstein, C.~E. Priebe, and
  G.~Sapiro, ``Graph matching: Relax at your own risk,'' {\em IEEE Transactions
  on Pattern Analysis and Machine Intelligence}, vol.~38, pp.~60--73, Jan 2016.

\bibitem{lylefipr16vn3}
V.~Lyzinski, K.~Levin, D.~E. Fishkind, and C.~E. Priebe, ``On the consistency
  of the likelihood maximization vertex nomination scheme: Bridging the gap
  between maximum likelihood estimation and graph matching,'' {\em Journal of
  Machine Learning Research}, vol.~17, no.~179, pp.~1--34, 2016.

\bibitem{burkard1997qaplib}
R.~E. Burkard, S.~E. Karisch, and F.~Rendl, ``Qaplib--a quadratic assignment
  problem library,'' {\em Journal of Global optimization}, vol.~10, no.~4,
  pp.~391--403, 1997.

\bibitem{zaslavskiy2009path}
M.~Zaslavskiy, F.~Bach, and J.-P. Vert, ``A path following algorithm for the
  graph matching problem,'' {\em IEEE Transactions on Pattern Analysis and
  Machine Intelligence}, vol.~31, no.~12, pp.~2227--2242, 2009.

\bibitem{liqixu12}
Z.-Y. Liu, H.~Qiao, and L.~Xu, ``An extended path following algorithm for
  graph-matching problem,'' {\em Pattern Analysis and Machine Intelligence,
  IEEE Transactions on}, vol.~34, no.~7, pp.~1451--1456, 2012.

\bibitem{Holland1983}
P.~W. Holland, K.~B. Laskey, and S.~Leinhardt, ``{Stochastic blockmodels: First
  steps},'' {\em Social Networks}, vol.~5, no.~2, pp.~109--137, 1983.

\bibitem{frwo56}
M.~Frank and P.~Wolfe, ``An algorithm for quadratic programming,'' {\em Naval
  Research Logistics Quarterly}, vol.~3, pp.~95--110, 1956.

\bibitem{ku55}
H.~W. Kuhn, ``The hungarian method for the assignment problem,'' {\em Naval
  Research Logistics Quarterly}, vol.~2, pp.~83--97, 1955.

\bibitem{budema2012}
R.~Burkard, M.~Dell'Amico, and S.~Marello, {\em Assignment Problems: Revised
  Reprint}.
\newblock Philadelphia: Sociaty for Industrial and Applied Mathematics (SIAM),
  2012.

\bibitem{coppersmith1987matrix}
D.~Coppersmith and S.~Winograd, ``Matrix multiplication via arithmetic
  progressions,'' {\em Journal of Symbolic Computation}, vol.~9, pp.~251--280,
  1990.

\bibitem{gurobi}
I.~Gurobi~Optimization, ``Gurobi optimizer reference manual,'' 2016.

\bibitem{enronwiki}
Wikipedia, ``Enron,'' {\em Wikipedia: The Free Encyclopedia}, 2017.
\newblock Online; accessed: 31-March-2017.

\bibitem{nytenron}
J.~Markoff, ``Armies of expensive lawyers, replaced by cheaper software,'' {\em
  The New York Times}, p.~A1, March 2011.

\bibitem{PCMP2005}
C.~E. Priebe, J.~M. Conroy, D.~J. Marchette, and Y.~Park, ``Scan statistics on
  enron graphs,'' {\em Computational \& Mathematical Organization Theory},
  vol.~11, no.~3, pp.~229--247, 2005.

\bibitem{lall2006genome}
S.~Lall, D.~Gr{\"u}n, A.~Krek, K.~Chen, Y.-L. Wang, C.~N. Dewey, P.~Sood,
  T.~Colombo, N.~Bray, P.~MacMenamin, {\em et~al.}, ``A genome-wide map of
  conserved micro{RNA} targets in c. elegans,'' {\em Current biology}, vol.~16,
  no.~5, pp.~460--471, 2006.

\bibitem{singh2008global}
R.~Singh, J.~Xu, and B.~Berger, ``Global alignment of multiple protein
  interaction networks with application to functional orthology detection,''
  {\em Proceedings of the National Academy of Sciences}, vol.~105, no.~35,
  pp.~12763--12768, 2008.

\bibitem{valouev2008high}
A.~Valouev, J.~Ichikawa, T.~Tonthat, J.~Stuart, S.~Ranade, H.~Peckham, K.~Zeng,
  J.~A. Malek, G.~Costa, K.~McKernan, {\em et~al.}, ``A high-resolution,
  nucleosome position map of c. elegans reveals a lack of universal
  sequence-dictated positioning,'' {\em Genome research}, vol.~18, no.~7,
  pp.~1051--1063, 2008.

\bibitem{celegans1}
L.~R. Varshney, B.~L. Chen, E.~Paniagua, D.~H. Hall, and D.~B. Chklovskii,
  ``Structural properties of the {C}aenorhabditis {E}legans neuronal network,''
  {\em PLoS Computational Biology}, vol.~7, no.~2, 2011.

\bibitem{bougleux2017graph}
S.~Bougleux, L.~Brun, V.~Carletti, P.~Foggia, B.~Ga{\"u}z{\`e}re, and M.~Vento,
  ``Graph edit distance as a quadratic assignment problem,'' {\em Pattern
  Recognition Letters}, vol.~87, pp.~38--46, 2017.

\end{thebibliography}

\end{document}